\documentclass{article}
\usepackage{amssymb}

\usepackage{mathtools} 
\usepackage{siunitx} 
\usepackage{amsmath}
\usepackage{amssymb}
\usepackage{booktabs}
\usepackage{tikz}
\usepackage{xcolor}
\usepackage{graphicx}
\usepackage{caption}
\usepackage{subcaption}
\usepackage{amsmath}
\usepackage{multirow} 

\definecolor{outlier}{RGB}{0,0,0}
\definecolor{car}{RGB}{255,153,0}
\definecolor{road}{RGB}{255,0,255}  
\definecolor{parking}{RGB}{255,153,204}
\definecolor{sidewalk}{RGB}{102,0,102}
\definecolor{building}{RGB}{0,204,255}
\definecolor{fence}{RGB}{51,102,255}
\definecolor{vegetation}{RGB}{0,128,0}
\definecolor{trunk}{RGB}{0,51,102}
\definecolor{terrain}{RGB}{204,255,204}
\definecolor{pole}{RGB}{204,255,255}
\definecolor{trafficsign}{RGB}{0,0,255}
\definecolor{CorrectClass}{RGB}{252,255,138}

\usepackage[preprint]{corl_2025} 

\title{Uncertainty-Aware Scene Understanding via Efficient Sampling-Free Confidence Estimation}

\author{
  Hanieh Shojaei Miandashti\\
  Institute of Cartography and Geoinformatics\\
  Leibniz University Hannover, 
  Germany\\
  \texttt{Hanieh.Shojaei@ikg.uni-hannover.de} \\
  \And
  Qianqian Zou \\
  Institute of Cartography and Geoinformatics\\
  Leibniz University Hannover, 
  Germany\\
  \texttt{qianqian.zou@ikg.uni-hannover.de} \\
  \And
  Claus Brenner \\
  Institute of Cartography and Geoinformatics\\
  Leibniz University Hannover, 
  Germany\\
  \texttt{claus.brenner@ikg.uni-hannover.de}
}

\begin{document}
\maketitle


\begin{abstract}

Reliable scene understanding requires not only accurate predictions but also well-calibrated confidence estimates to ensure calibrated uncertainty estimation, especially in safety-critical domains like autonomous driving. In this context, semantic segmentation of LiDAR points supports real-time 3D scene understanding, where reliable uncertainty estimates help identify potentially erroneous predictions. While most existing calibration approaches focus on modeling epistemic uncertainty, they often overlook aleatoric uncertainty arising from measurement inaccuracies, which is especially prevalent in LiDAR data and essential for real-world deployment.
In this work, we introduce a sampling-free approach for estimating well-calibrated confidence values by explicitly modeling aleatoric uncertainty in semantic segmentation, achieving alignment with true classification accuracy and reducing inference time compared to sampling-based methods. Evaluated on the real-world SemanticKITTI benchmark, our approach achieves 1.70\% and 1.33\% Adaptive Calibration Error (ACE) in semantic segmentation of LiDAR data using RangeViT and SalsaNext models, and is more than one order of magnitude faster than the comparable baseline. Furthermore, reliability diagrams reveal that our method produces underconfident rather than overconfident predictions — an advantageous property in safety-critical systems.

\end{abstract}



\section{Introduction}

In safety-critical domains such as autonomous driving, deep neural networks (DNNs) must ensure both high accuracy and well-calibrated confidence estimates to support reliable uncertainty estimation, allowing the system to recognize when it is likely to be wrong and act conservatively. For 3D scene understanding from LiDAR point clouds, this means identifying points where the predicted semantic labels may be unreliable. Ideally, confidence values should align with the true likelihood of correctness; however, modern DNNs often produce overconfident outputs, failing to capture inherent uncertainties in data and model predictions \citep{guo2017calibration, ovadia2019can, wang2021rethinking}. Such miscalibration is especially concerning in safety-critical settings, where high-confidence errors can lead to unsafe decisions. Proper calibration is thus essential for building trustworthy autonomous systems.

A variety of approaches have been proposed to improve confidence calibration, including post-hoc techniques and uncertainty quantification methods \citep{gawlikowski2023survey, wang2023calibration}. Among post-hoc methods, temperature scaling is widely used due to its simplicity and effectiveness in calibrating DNNs \citep{guo2017calibration}. However, uncertainty-aware approaches such as deep ensembles (DE) \citep{lakshminarayanan2017simple} and Monte Carlo Dropout (MC dropout) \citep{gal2016dropout} have shown better calibration performance \citep{ovadia2019can}. Recent studies have further shown that temperature scaling becomes ineffective when class distributions overlap significantly, particularly as the number of classes increases \citep{chidambaram2023limitations}. To address this limitation, we propose a method that explicitly models aleatoric uncertainty by representing each class logit as a Gaussian distribution and incorporating distributional overlap into confidence estimation. Confidence is estimated as the probability that the predicted class yields a higher sampled logit than all competing classes, effectively quantifying the likelihood of a correct prediction. While this probability is typically approximated via inefficient Monte Carlo sampling, we introduce a closed-form lower bound that eliminates the need for sampling in multi-class settings, making the approach well-suited for real-time applications.  

Reliable confidence calibration must account for both aleatoric uncertainty and epistemic uncertainty. Aleatoric uncertainty arises from irreducible inherent noise in the data, such as LiDAR sensor inaccuracies, varying point cloud density with distance, environmental variability, and surface reflectivity. In contrast, epistemic uncertainty stems from limited model knowledge and can be reduced with more data \citep{gal2016uncertainty}. While most existing methods focus primarily on epistemic uncertainty \citep{lakshminarayanan2017simple, gal2016dropout, mukhoti2023deep, osband2023epistemic}, our approach further incorporates aleatoric uncertainty directly into the confidence estimation process, leading to better calibration on complex, noisy input data like LiDAR point clouds. I.e., we integrate epistemic uncertainty with our aleatoric uncertainty to produce well-calibrated confidence estimates.

Our main contribution is a novel confidence estimation method that accounts for the overlap between logit distributions when computing confidence values. We make three key claims. \textbf{First}, our sampling-free approach generates confidence values that closely approximate true confidence levels, effectively calibrated against the actual classification accuracy. Additionally, our method produces underconfident rather than overconfident values, making it particularly valuable for safety-critical decision-making. \textbf{Second}, our confidence estimation approach outperforms the comparable approaches (e.g., temperature scaling) by accounting for the underlying data distribution. Moreover, when combined with epistemic uncertainty, it achieves the highest calibration performance. \textbf{Third}, our proposed sampling-free approach reduces inference time compared to sampling methods such as logit-sampling while maintaining confidence calibration and classification accuracy. These proposed contributions are validated through experiments conducted on benchmark datasets of SemanticKITTI \citep{behley2019semantickitti} and nuScenes \citep{caesar2020nuscenes}, evidenced by our performance on the Adaptive Calibration Error (ACE) \citep{nixon2019measuring} metric and further observed in the reliability diagram \citep{guo2017calibration}.


\section{Related Works}
\label{sec:related works}

\subsection{Confidence calibration}
\label{sec:Confidence Calibration_RelatedWorks}

Confidence values were first introduced as Maximum Class Probability (MCP), the highest probability in the softmax distribution, based on the assumption that correctly classified samples generally have higher MCP than misclassified and out-of-distribution examples \citep{hendrycks2016baseline}. However, subsequent studies have identified major limitations in MCP: it often produces overly confident estimates, highlighting the need for confidence calibration \citep{jiang2018trust, corbiere2019addressing, guo2017calibration, thulasidasan2019mixup}. 

Previous research on confidence calibration of deep learning models typically falls into two categories. The first involves post-hoc methods that adjust classifier outputs by rescaling the logits without retraining the model \citep{guo2017calibration, kull2019beyond, zadrozny2002transforming}. Temperature scaling \citep{guo2017calibration} exemplifies this by recalibrating model logits using a single parameter optimized on a validation set.

While temperature scaling and its variants are widely used for post-hoc calibration, they primarily adjust predicted probabilities without addressing the underlying sources of miscalibration, such as model uncertainty or data noise. These methods do not modify the model itself and therefore fail to correct overconfidence that stems from overlapping or noisy class distributions. Recent work by \citet{chidambaram2023limitations} has shown, both theoretically and empirically, that temperature scaling becomes increasingly ineffective as class overlap grows, and can asymptotically perform no better than random guessing in multi-class settings. In contrast, the second category incorporates uncertainty directly into the training phase, enhancing models' inherent ability to account for data variability. 

\subsection{Uncertainty estimation}
\label{sec:Uncertainty Estimation}

Methods such as Bayesian neural networks (BNNs) \citep{neal2012bayesian} and evidential deep learning (EDL) \citep{sensoy2018evidential} equip models to inherently represent uncertainty, providing a more fundamental solution to confidence calibration challenges. While EDL calculates total uncertainty without differentiating between epistemic and aleatoric, BNNs specifically model epistemic uncertainty by placing a prior distribution over parameters of a model and approximating the posterior distribution through Bayesian inference~\citep{neal2012bayesian, louizos2017multiplicative}. However, due to the often intractable nature of exact inference, variational methods such as Bayes by backprop \citep{blundell2015weight} employ an evidence-based lower bound for approximating the posterior distribution. Several tractable methods for estimating epistemic uncertainty have emerged in recent years \citep{kendall2015bayesian, lee2017training, osband2023epistemic}, including MC dropout which applies dropout during inference and deep ensembles, which approximate the posterior distribution using an ensemble of networks with varied initializations. However, although these models effectively estimate confidence by averaging softmax outputs from multiple instances, they may still require additional calibration to better align softmax probabilities. This limitation arises because they primarily rely on the model’s softmax values without explicitly accounting for aleatoric uncertainty.

To accurately calibrate softmax outputs by accounting for the underlying true distribution, the logit-sampling approach proposed a method that assumes a Gaussian distribution for the logits of each class \citep{kendall2017uncertainties}. However, during its inference, the need to perform Monte Carlo sampling across each distribution to compute calibrated confidence introduces additional computational overhead and increases inference time. Our proposed approach lies in this line of work and quantifies aleatoric uncertainty from Gaussian logit distributions without sampling, making it suitable for real-time applications with many classes.

\subsection{Semantic segmentation of LiDAR point clouds}
3D scene perception using deep learning-based semantic segmentation of LiDAR point clouds can be classified into two categories based on their underlying 3D representations \citep{guo2020deep}: The first category includes point-wise methods that process 3D data directly, including raw 3D point-based architectures \citep{qi2017pointnet++, hu2020randla, puy2023using, thomas2019kpconv} and voxel-based networks \citep{choy20194d, cheng20212, tang2020searching, zhu2021cylindrical, huang2016point}. While voxel-based methods convert unordered point clouds into structured 3D grids—enabling the use of 3D convolutions to capture geometric features—they are often computationally intensive. In contrast, the second category comprises projection-based methods, which convert 3D point clouds into 2D representations, either as bird’s-eye view maps \citep{chen2021polarstream, zhang2020polarnet} or spherical range-view images (panoramic view) \citep{cortinhal2020salsanext, xu2020squeezesegv3, milioto2019rangenet++, kong2023rethinking, ando2023rangevit}. These methods benefit from the structured 2D image representations, enabling the use of 2D convolutional neural networks (CNNs) and vision transformers (ViTs), which are more computationally efficient.

In this work, we use SalsaNext, a CNN model, and RangeViT, a transformer-based model that achieves state-of-the-art performance in LiDAR semantic segmentation on benchmark datasets of SemanticKITTI \citep{behley2019semantickitti} and nuScenes \citep{caesar2020nuscenes} using 2D range-view representations.

\section{Methodology}
\label{sec:Methodology}

Our approach predicts Gaussian distributions over the logit outputs of a classification model by directly estimating the mean ($\boldsymbol{\mu}_k$) and variance ($\boldsymbol{\sigma^2}_k$) for each $k$-th input before applying the softmax function. During training, considering $T$ samples from each Gaussian distribution over $C$ classes, the loss function is calculated based on the average of the softmax results from the sampled logits, rather than from a single logit score.

Considering the class with the highest mean as the predicted class, we directly compute its well-calibrated confidence value by evaluating the probability that logit scores from this class exceed those of competing classes, thus efficiently capturing aleatoric uncertainty without the need for sampling from Gaussian distributions.

\subsection{Confidence computation}
\label{sec:Confidence Computation}

\subsubsection{Exact and approximate computation}

The proposed method is based on the Gaussian distributions over the logits.
As usual, one would predict the class whose predicted mean is maximal. Let there be $C$ classes, for each of which a Gaussian distribution $\mathcal{N}(\cdot | \mu_i, \sigma_i^2)$, $1\leq i\leq C$ is predicted. Then, without loss of generality, we may assume $\mu_1 \geq \mu_i$ for $i\geq 2$, so that class $1$ is selected as the predicted class. The confidence is then defined as the probability that this selection is correct, which, given random variables $X_i \sim \mathcal{N}(\cdot | \mu_i, \sigma_i^2)$, is $P(X_1 \geq \{X_i\}_{i\geq 2})$, equivalent to $P(X_1 \geq \max_{i\geq 2} X_i)$.

As there is no closed-form solution to compute this probability in the general case, it may be approximated by computing a relative frequency by simulation. Similar to the training phase, $C$ samples $X_i$, $1\leq i\leq C$ are drawn (one from each Gaussian), and it is determined if $X_1$ is largest. From repeating this experiment $N$ times, the relative frequency of cases $X_1 \geq \max_{i\geq 2}X_i$ is computed. This method requires $NC$ draws.

It is easily seen that the required probability is given by 

\begin{equation}
\label{eq:McIntegral}
P(X_1 \geq \max_{i\geq 2}X_i) =
\int_{-\infty}^{+\infty} \varphi_1(x)\prod_{i=2}^C \Phi_i(x) dx,
\end{equation}
where we have used
$\varphi_i(x)=\mathcal{N}(x | \mu_i, \sigma_i^2)$ for the Gaussian densities and $\Phi_i(x)$ for their associated cumulative distribution functions. This integral can be approximated via Monte Carlo simulation using
\begin{equation}
\label{eq:MC-integration}
P(X_1 \geq \max_{i\geq 2}X_i) \approx
\frac{1}{N}\sum_{k=1}^N \prod_{i=2}^C \Phi_i(x_k) ,
\end{equation}
with $x_k\sim\varphi_1$, requiring only $N$ samples to be drawn (which in fact can be re-used by scaling and shifting a fixed set of samples).

\subsubsection{Sampling-free confidence quantification}
\label{sec:Sampling-free Confidence Value Quantification}

For the special case of two classes, a closed-form solution can be given, because $P(X_1\geq X_2) = P(Z\geq 0)$ with $Z:=X_1-X_2$. As $Z\sim\mathcal{N}(\cdot | \mu_1-\mu_2, \sigma_1^2+\sigma_2^2)$ it follows that
\begin{equation}
\label{eq:confidence12}
P(X_1\geq X_2) =
\Phi(\mu_1-\mu_2 | 0, \sigma_1^2+\sigma_2^2) =: \Phi_{1,2},
\end{equation}
which unfortunately does not extend readily to $C>2$ classes. However, using pairwise $\Phi_{1,i}$, $i\geq 2$, the following lower bound holds
\begin{equation}
\label{eq:confidence}
P(X_1 \geq \max_{i\geq 2}X_i) \geq
\prod_{i=2}^C \Phi_{1,i},
\end{equation}
which follows from the fact that the integral in Equation~\ref{eq:McIntegral} is the expected value
$\mathbb{E}[\prod_{i=2}^C \Phi_i(X)]$
under the distribution ${X\sim \varphi_1}$, and since all
$\Phi_i(x)$ are strictly monotonically increasing functions of $x$, their covariance is non-negative, from which it follows that
$\mathbb{E}[\Phi_i(X)\Phi_j(X)]\geq
\mathbb{E}[\Phi_i(X)]\mathbb{E}[\Phi_j(X)]$, yielding Equation~\ref{eq:confidence}. For products of more than two terms, this follows by recursive application, noting that any product of Gaussian cumulative distribution functions is strictly monotonically increasing as well. A detailed derivation of Equation~\ref{eq:confidence} is provided in Section~\ref{sec:derivationLowerBound}.

To conclude, given the predicted class distributions, a lower bound for the confidence can be computed by simply evaluating $C-1$ `pairwise' cumulative distribution functions, requiring no sampling. To give some intuition, if class $1$ is a clear winner, the confidence will be 1. If the winner class is challenged only by one alternative class, the lower bound Equation~\ref{eq:confidence} reduces to Equation~\ref{eq:confidence12}, and the bound is exact. If more than one other class challenges the winner class, the lower bound will underestimate the confidence.

In our experiments, the proposed lower bound exhibits only a negligible difference when compared to the exact formulation. As demonstrated in Section~\ref{sec:Exact vs. LB}, this lower bound provides a close approximation to the true value obtained through Monte Carlo integration, while exhibiting a slightly underconfident behavior.

\section{Experiments}
\label{sec:Experiments}

In this section, we compare confidence values from our lower bound approach with exact and sampling-based methods, as detailed in Section~\ref{sec:Exact vs. LB}, to support our first claim of achieving comparable calibration without sampling. Section~\ref{sec:Confidence calibration analysis} evaluates the calibration performance of our method against temperature scaling and shows further improvements when combined with epistemic uncertainty, supporting our second claim. Section~\ref{sec:Inference time analysis} demonstrates the faster inference of our approach, validating our third claim. Additionally, we provide visualizations to illustrate uncertainty-aware LiDAR semantic segmentation in Section~\ref{sec:LiDAR qualitative results}.

\subsection{Experimental setup}
\label{sec:Experimental Details}

\subsubsection{Datasets and network architectures} 

We evaluate our approach on two widely used LiDAR semantic segmentation benchmarks—SemanticKITTI \citep{behley2019semantickitti} and nuScenes \citep{caesar2020nuscenes}. Each 3D scan from SemanticKITTI and nuScenes was converted based on their LiDAR beams into a $[64 \times 2048 \times 5]$ and $[32 \times 2048 \times 5]$ spherical range-view image, respectively. Each image contains five channels corresponding to 3D point coordinates $(x, y, z)$, intensity, and range values, serving as input for semantic segmentation. 

We employ SalsaNext \citep{cortinhal2020salsanext} as a CNN model which adopts a U-Net encoder-decoder architecture enhanced with ResNet blocks for efficient feature extraction, and RangeViT, a transformer-based model that exploits Vision Transformers (ViTs) and achieves state-of-the-art performance in LiDAR semantic segmentation to validate our approach on more recent architectures. We employ a composite loss function combining the multi-class focal loss \citep{lin2017focal} and the Lovász-Softmax loss \citep{berman2018lovasz}, aiming to enhance both confidence calibration and segmentation performance.

\subsubsection{Comparative methods} 
\label{sec:Comparative Methods}
To evaluate our proposed method, we use MCP as the baseline uncalibrated confidence measure, and apply temperature scaling, logit-sampling, and our approach for confidence calibration. Each method is further combined with epistemic uncertainty modeling using deep ensembles and MC dropout. Additionally, we assess EDL as a competitive calibration strategy, which estimates predictive uncertainty without explicitly separating aleatoric and epistemic components. The evaluation of all methods is summarized in Table~\ref{tab:time_ace_semantickitti_nuscenes} of Section~\ref{sec:Exact vs. LB}.

\subsubsection{Evaluation metrics}
\label{sec:Evaluation Metrics}
The classification performance is commonly evaluated using the mIoU metric, which accounts for the inherent class imbalance in semantic segmentation datasets and is well-suited for real-world applications. To evaluate the calibration of the confidence values for predicted classes, we use the ACE \citep{nixon2019measuring}, complemented by reliability diagrams \citep{guo2017calibration}, which illustrate whether the model is underconfident or overconfident—information that ACE does not provide, as it only quantifies the absolute deviation from perfect calibration. In contrast to the Expected Calibration Error (ECE) \citep{naeini2015obtaining}, ACE assigns equal weight to each bin in the reliability diagram, defined as $\text{ACE} = \frac{1}{M} \sum_{m=1}^{M} |c_m - \text{Acc}_m|$, where \( M \) represents the number of non-empty bins, \( c_m \) is the average confidence within bin \( m \), and \( \text{Acc}_m \) is the corresponding average accuracy.

\subsection{Comparative analysis of sampling-based and sampling-free confidence computations} 
\label{sec:Exact vs. LB}  
\begin{figure}
\begin{minipage}{.485\textwidth}
\centering
\subfloat[Ours vs. exact\label{fig:lb vs exact_a}]
{\includegraphics[width=.45\linewidth]{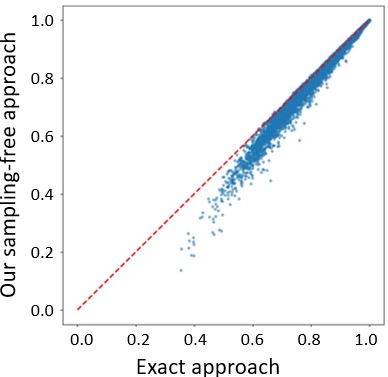}}\quad
\subfloat[Reliability diagram \label{fig:lb vs exact_b}]{\includegraphics[width=.45\linewidth]{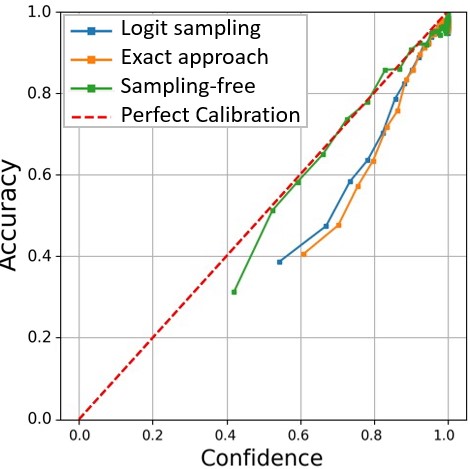}}
\caption{Comparison of confidence estimation methods: (a) Scatter plot shows minimal difference between exact and sampling-free confidences; (b) Reliability diagram indicates our method produces slightly underconfident predictions compared to baselines.}
\label{fig:lb vs exact}
\end{minipage}\hfill
\begin{minipage}{.485\textwidth}
\centering
\subfloat[SalsaNext \label{fig:salsanext_calib}]{\includegraphics[width=.45 \linewidth, trim={0.3cm 0.3cm 0.30cm 0.8cm},clip]{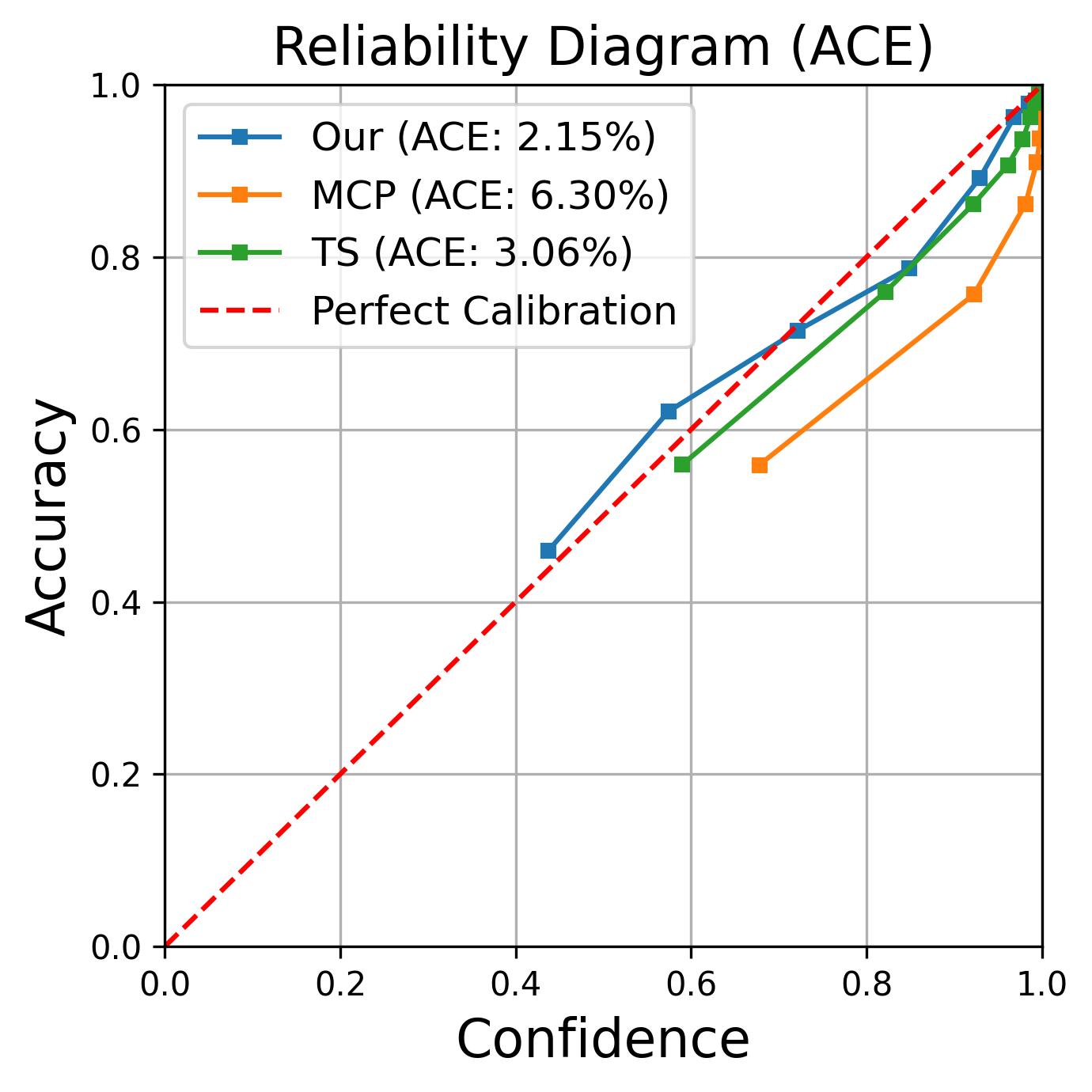}} \quad
\subfloat[RangeViT  \label{fig:rangevit_calib}]{\includegraphics[width=.45 \linewidth, trim={0.3cm 0.3cm 0.30cm 0.8cm},clip]{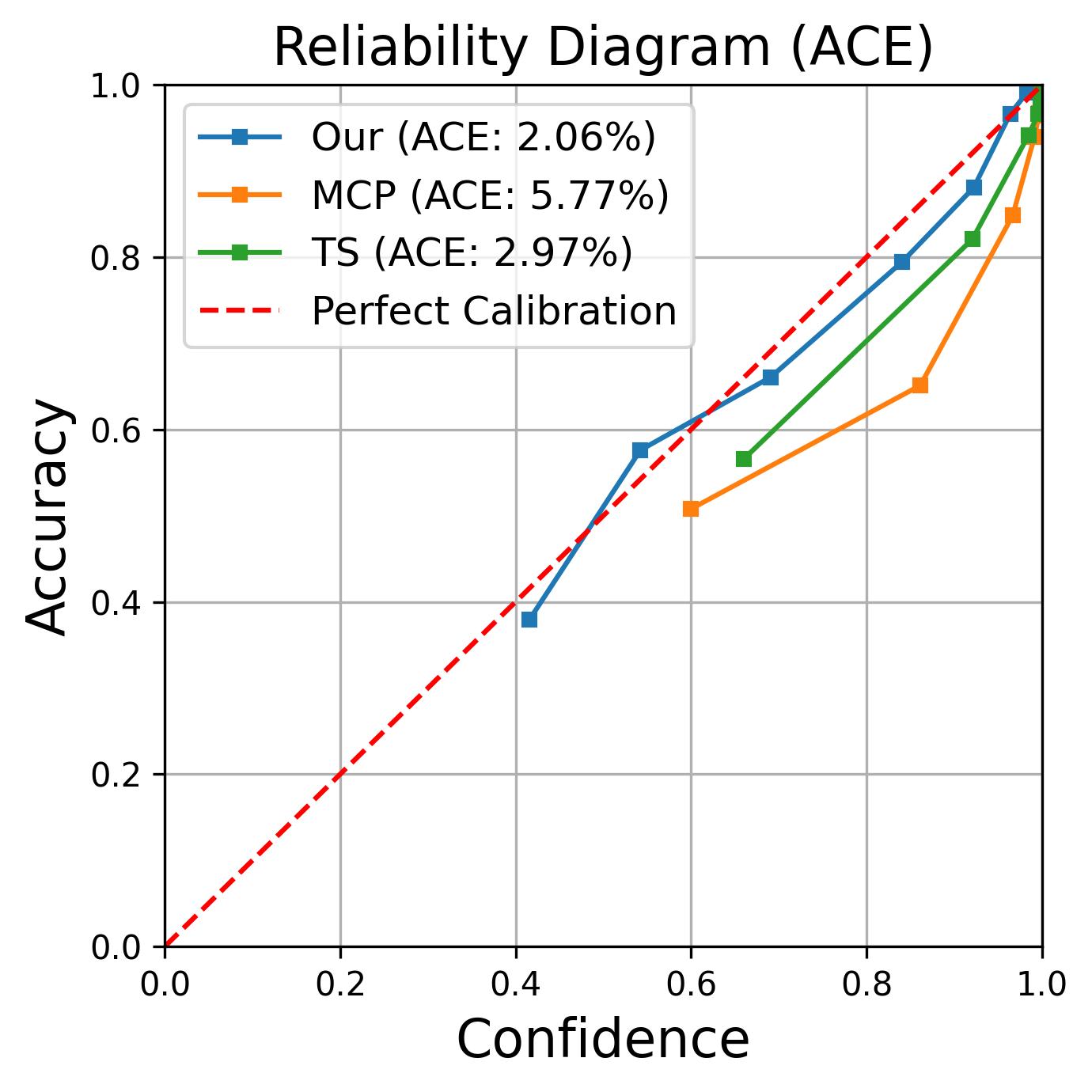}}
\caption{Reliability diagrams comparing calibration of our method against temperature scaling (TS) and uncalibrated (MCP) models on SemanticKITTI validation set using SalsaNext and RangeViT. Our method shows better calibration (closer to perfect calibration).}
\label{fig:calib}
\end{minipage}
\end{figure}

To support our first claim that the sampling-free lower bound approach closely estimates the true confidence values, we compare these confidence values with those obtained from the Monte Carlo integration. They are computed based on Equation~\ref{eq:confidence} and Equation~\ref{eq:MC-integration}, respectively. The scatter plot in Figure~\ref{fig:lb vs exact_a} illustrates this comparison on a subset of test samples of SemanticKITTI with SalsaNext model, with $x$-axis representing the exact confidence values and the $y$-axis showing the lower bound estimation. The red dashed line denotes the ideal ${y=x}$ line, where the lower bound would match the exact computation. 

Figure~\ref{fig:lb vs exact_a} shows that the lower bound estimation predominantly aligns closely with or slightly underestimates the exact values, evidenced by the clustering of points below the \(y=x\) line. This pattern indicates that the lower bound estimation tends to behave conservatively, often yielding slightly underestimated confidence values across a broad range. 

Additionally, Figure~\ref{fig:lb vs exact_b} contrasts the true classification accuracy ($y$-axis) with the exact confidence computation, the sampling-free lower bound approach and the logit-sampling baseline. Here, the sampling-free lower bound approach, depicted by the green line, aligns more closely with the ideal red line and consistently shows more conservative confidence estimates relative to the logit-sampling baseline and the exact values. This visualization further highlights the conservative nature of our sampling-free approach.

\begin{figure}[ht!]
\centering

\begin{minipage}[t]{0.48\textwidth}
    \centering
    \begin{subfigure}[t]{\linewidth}
        \includegraphics[width=\linewidth]{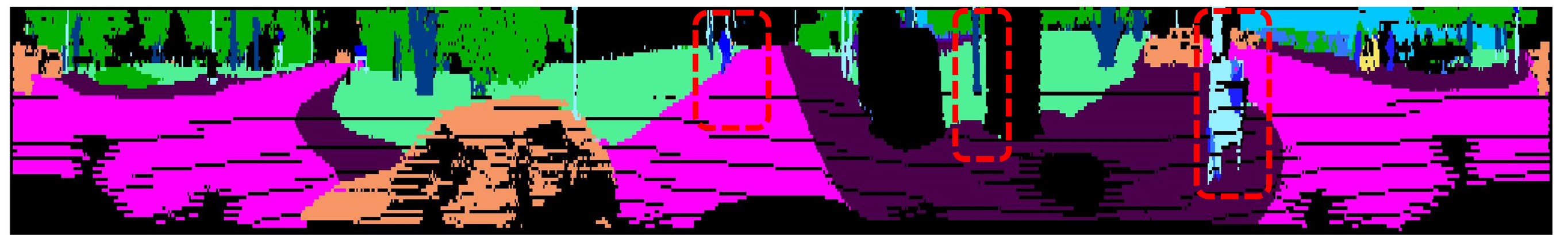}
        \caption{Our prediction map.}
        \label{fig:79_pred}
    \end{subfigure}
    \vspace{0.2cm}
    \begin{subfigure}[t]{\linewidth}
        \includegraphics[width=\linewidth]{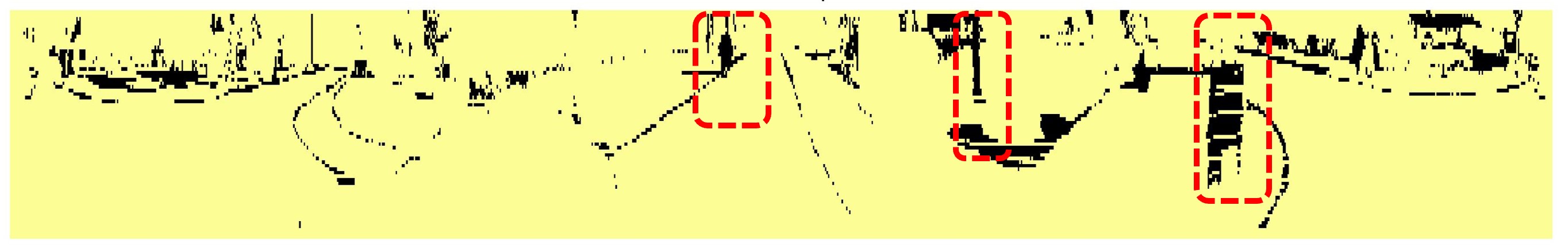}
        \caption{Our error map.}
        \label{fig:79_error}

    \end{subfigure}
    \vspace{0.2cm}
    \begin{subfigure}[t]{\linewidth}
        \includegraphics[width=\linewidth]{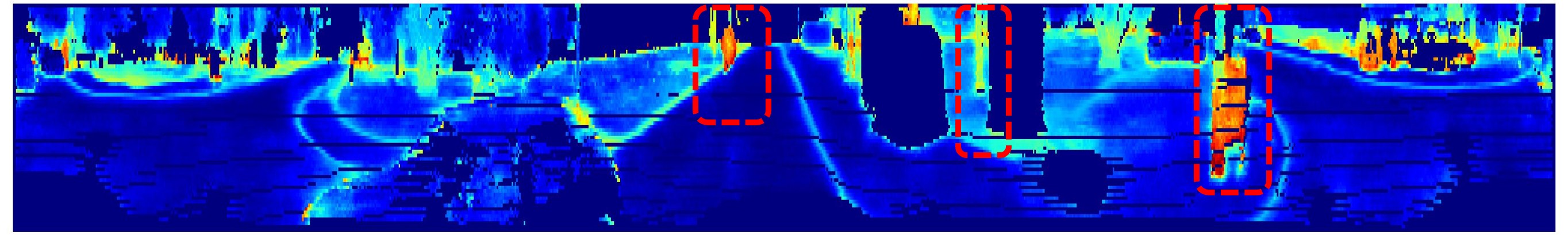}
        \caption{our uncertainty map.}
         \label{fig:79_our}
    \end{subfigure}
    \vspace{0.2cm}
    \begin{subfigure}[t]{\linewidth}
        \includegraphics[width=\linewidth]{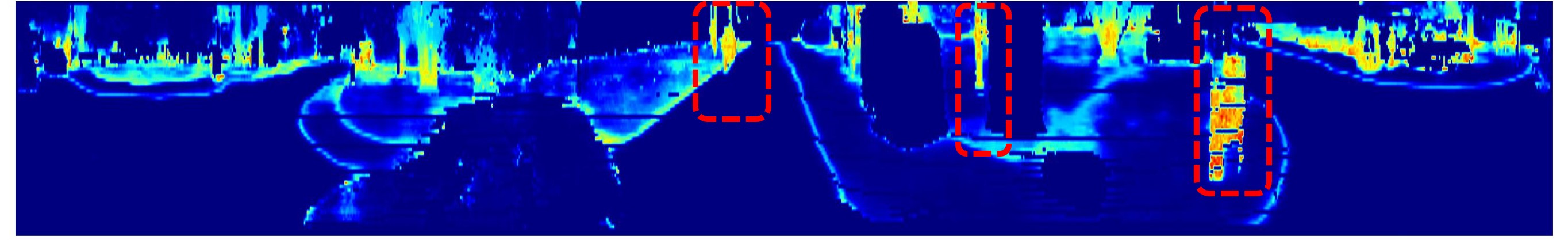}
        \caption{Logit-sampling uncertainty map.}
         \label{fig:79_ls}
    \end{subfigure}
    \vspace{0.2cm}
     \includegraphics[width=\linewidth]{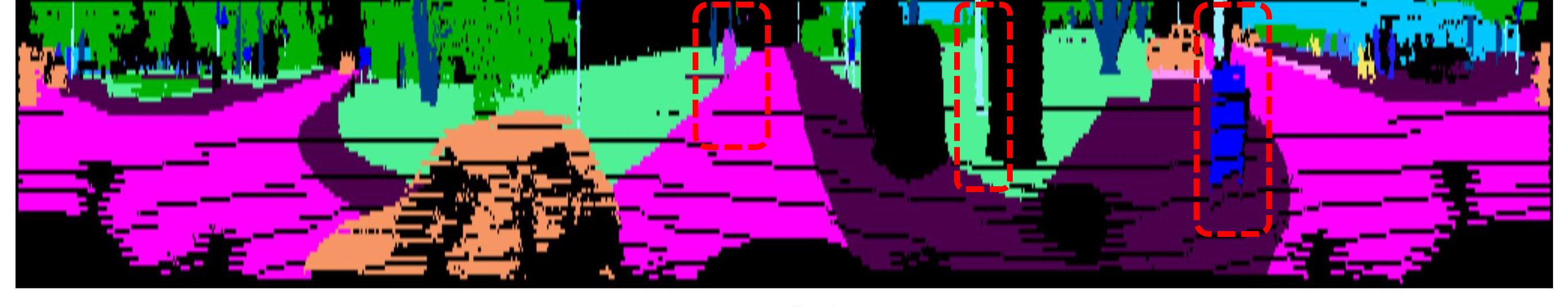}
    \par\vspace{0.1cm}{\small \centering Ground truth.\par}
    \vspace{0.1cm}

\end{minipage}
\hfill
\begin{minipage}[t]{0.48\textwidth}
    \centering
    \begin{subfigure}[t]{0.97\linewidth}
        \includegraphics[width=\linewidth]{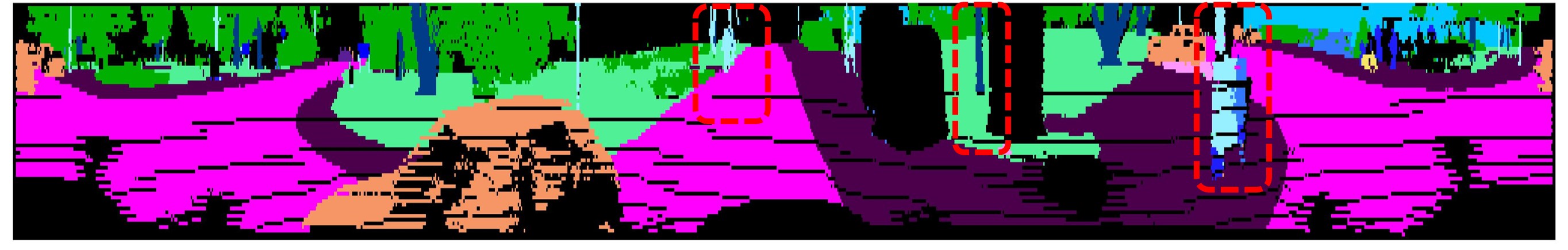} 
        \caption{Temperature scaling prediction map.}
    \end{subfigure}
    \vspace{0.2cm}
    \begin{subfigure}[t]{\linewidth}
        \includegraphics[width=\linewidth]{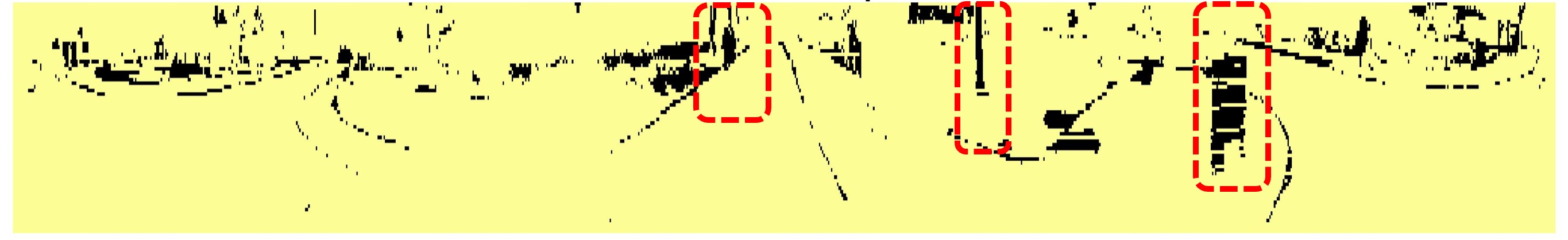}  
        \caption{Temperature scaling error map.}
    \end{subfigure}
    \vspace{0.2cm}
    \begin{subfigure}[t]{\linewidth}
        \includegraphics[width=\linewidth]{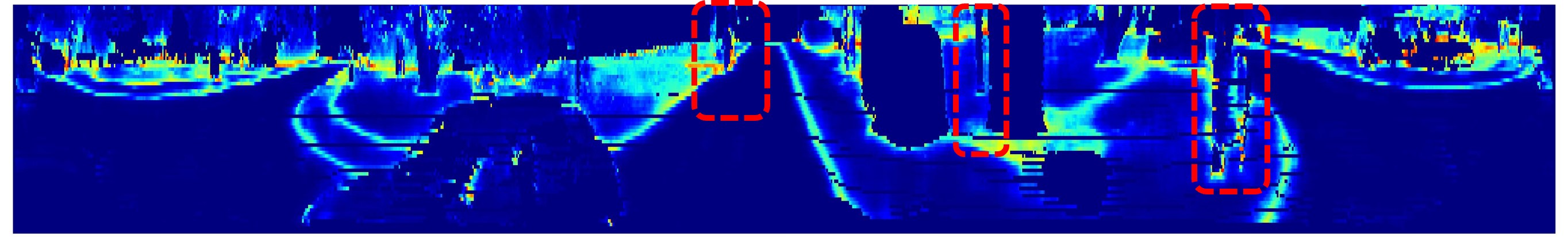} 
        \caption{Temperature scaling uncertainty map.}
        \label{fig:79_ts}
    \end{subfigure}
    \vspace{0.2cm}
    \begin{subfigure}[t]{\linewidth}
        \includegraphics[width=\linewidth]{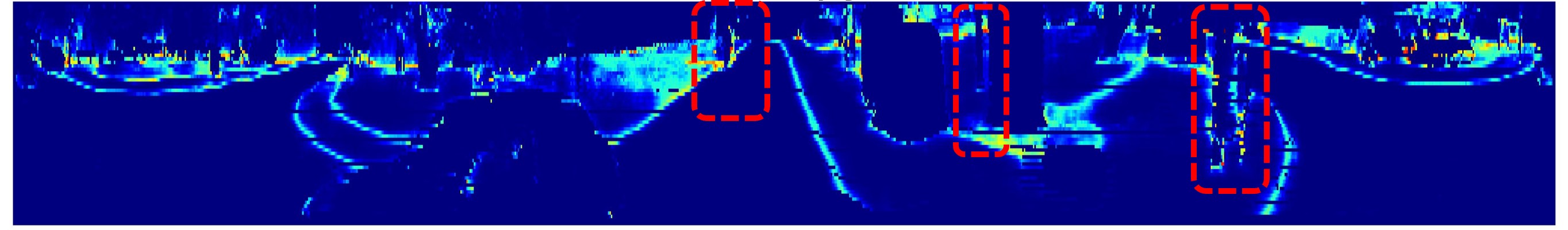} 
        \caption{MCP uncertainty map.}
        \label{fig:79_mcp}
    \end{subfigure}
    \vspace{0.2cm}
     \includegraphics[width=0.8\linewidth]{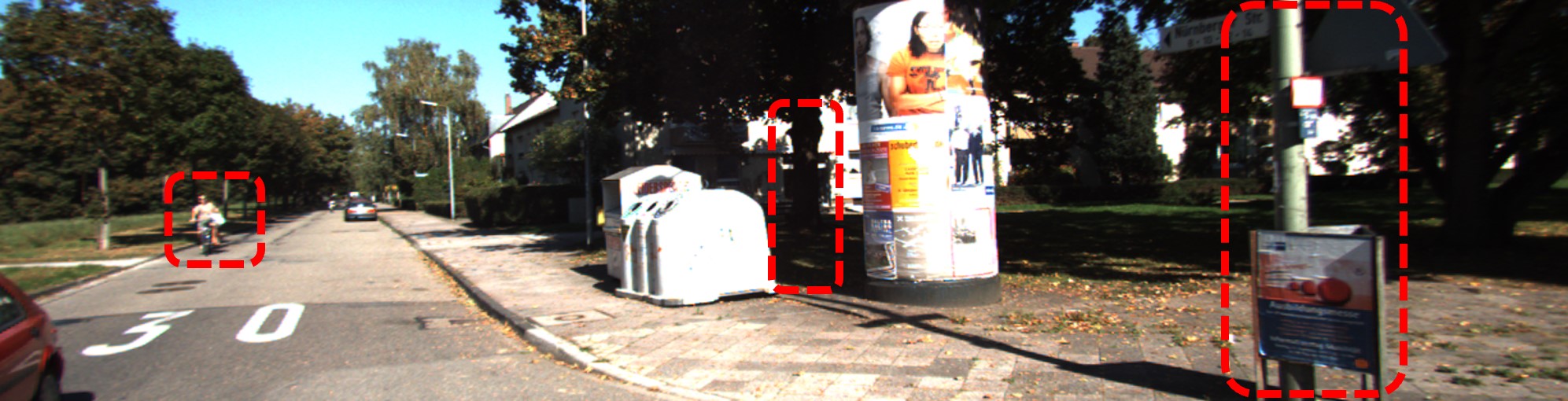}
    \par\vspace{0.2cm}{\small \centering Camera image. \par}
    \vspace{0.2cm}
\end{minipage}

\caption{Qualitative comparison of uncertainty maps from the logit-sampling baseline and our sampling-free method (a-d). Both align with misclassifications, but our method shows higher uncertainty in error-prone regions, reflecting more underconfident estimation. Uncertainty maps are compared to those from temperature scaling and an uncalibrated model (e-h), which has not detected those misclassifications as high-uncertainty regions. \colorbox{outlier}{\textcolor{white}{outlier}},\colorbox{car}{\textcolor{white} {car}},\colorbox{road}{\textcolor{white}{road}},\colorbox{sidewalk}{\textcolor{white}{sidewalk}},\colorbox{building}{\textcolor{white}{building}},\colorbox{fence}{\textcolor{white}{fence}},\colorbox{vegetation}{\textcolor{white}{vegetation}},\colorbox{trunk}{\textcolor{white}{trunk}},\colorbox{terrain}{terrain},\colorbox{pole}{pole},\colorbox{trafficsign}{\textcolor{white}{traffic-sign}}. }
\label{fig:Results_scan79}
\end{figure}

To further illustrate the behavior of our sampling-free confidence estimation, Figure~\ref{fig:Results_scan79} compares uncertainty maps from the logit-sampling baseline (\ref{fig:79_ls}) and our method (\ref{fig:79_our}) on a representative SemanticKITTI scan. Both methods exhibit similar patterns aligned with model misclassification (error map shown in \ref{fig:79_error}), validating the correctness of our estimation. However, our method assigns higher uncertainty in erroneous regions—such as misclassified bike, trunk, and pole instances—highlighted by red dashed boxes, demonstrating its more conservative (underconfident) behavior.


\subsection{Confidence calibration analysis} 
\label{sec:Confidence calibration analysis}

In this section, we provide experimental evidence supporting our second claim that our approach produces better-calibrated confidence estimates than temperature scaling. Figure~\ref{fig:calib} presents reliability diagrams of our confidence calibration method across two backbone networks—SalsaNext (\ref{fig:salsanext_calib}) and RangeViT (\ref{fig:rangevit_calib})—on the SemanticKITTI dataset, alongside the uncalibrated model (MCP) and the temperature-scaled variant (TS). While temperature scaling improves the calibration of MCP, it still deviates noticeably from the perfect calibration (red dashed line). In contrast, our method consistently remains closer to perfect calibration and displays mild underconfidence in the $0.4$ to $0.7$ confidence range. On SalsaNext, our approach achieves an ACE of $2.15\%$, outperforming TS ($3.06\%$) and MCP ($6.30\%$). Similarly, for RangeViT, our method attains an ACE of $2.06\%$, compared to $2.97\%$ for TS and $5.77\%$ for MCP. These results confirm the superior calibration performance of our approach across both architectures. Qualitative results in Figure~\ref{fig:Results_scan79} support our findings: three misclassified objects exhibit low uncertainty in the maps produced by the uncalibrated model (\ref{fig:79_mcp}) and temperature scaling (\ref{fig:79_ts}), whereas our proposed method (\ref{fig:79_our}) assigns higher uncertainty to these samples and demonstrates improved accuracy, as evidenced by the comparison of the two error maps.   

To support our second claim—that combining epistemic uncertainty with aleatoric uncertainty achieves the highest calibration performance—we conduct further analysis by combining our aleatoric confidence estimation with two epistemic modeling approaches: DE \citep{lakshminarayanan2017simple} and MC dropout \citep{gal2016dropout}. Results in Table~\ref{tab:time_ace_semantickitti_nuscenes} demonstrate that once the epistemic uncertainty is incorporated, both our proposed approach and the logit-sampling method (which consider aleatoric in confidence estimation), paired with either DE or MC dropout, consistently outperformed all other methods in terms of ACE, achieving the lowest calibration error across both semantic segmentation networks ($1.70\%$ and $1.21\%$ on RangeViT and SalsaNext for SemanticKITTI and $1.78\%$ on RangeViT for nuScenes). These results highlight the significance of jointly modeling both aleatoric and epistemic uncertainty for effective confidence calibration.  

Overall, Table~\ref{tab:time_ace_semantickitti_nuscenes} shows that the lowest calibration errors are achieved by methods that incorporate aleatoric uncertainty, with further improvements when epistemic uncertainty is jointly modeled—whether via DE or MC dropout. These approaches consistently outperform EDL, temperature scaling and uncalibrated baselines. Notably, for both datasets with RangeViT, the best ACE ($1.70\%$ and $1.73\%$) is achieved by our sampling-free method combined with DE, a trend that also holds for SalsaNext, where the same combination yields an ACE of $1.33\%$. Qualitative results detailed in Section~\ref{sec:LiDAR qualitative results} demonstrate that our proposed approach, combined with DE, estimates predictive uncertainty that closely follows the error map, thereby producing uncertainty-aware semantic segmentation of LiDAR scans.

\subsection{Inference time analysis} 
\label{sec:Inference time analysis}
To validate our third claim—that our proposed confidence estimation method reduces inference time compared to sampling-based approaches for modeling aleatoric uncertainty—Table~\ref{tab:flops_time} compares inference time and FLOPs between our sampling-free method and the logit-sampling approach. As shown, our method achieves a substantial reduction in inference time, decreasing it by a factor of 15 times for RangeViT and 18 times for SalsaNext, while incurring only minimal computational overhead. Specifically, the logit-sampling method adds 6.82G FLOPs to both models due to 50 times sampling per pixel, whereas our method increases the total FLOPs by just 0.07G FLOPs through pairwise CDF computations. All results are measured on a GeForce RTX 3060 GPU.

\begin{table}[ht!]
\centering
\caption{Comparative analysis of inference time, mIoU, and ACE across confidence calibration methods on the SemanticKITTI and nuScenes validation sets using SalsaNext and RangeViT. The best-performing results are highlighted in bold, and the second-best are shown in blue.}
\label{tab:time_ace_semantickitti_nuscenes}
\resizebox{\linewidth}{!}{%
\begin{tabular}{@{}lcc|ccc|ccc|ccc@{}}
\toprule
\multirow{2}{*}{Method} & \multicolumn{2}{c|}{Uncertainty Type} & \multicolumn{3}{c|}{RangeViT (SemanticKITTI)} & \multicolumn{3}{c|}{SalsaNext (SemanticKITTI)} & \multicolumn{3}{c}{RangeViT (nuScenes)} \\
\cmidrule(lr){2-3} \cmidrule(lr){4-6} \cmidrule(lr){7-9} \cmidrule(lr){10-12}
& Aleatoric & Epistemic & mIoU (\%) $\uparrow$ & ACE (\%) $\downarrow$ & Time (ms) $\downarrow$ & mIoU (\%) $\uparrow$ & ACE (\%) $\downarrow$ & Time (ms) $\downarrow$ & mIoU (\%) $\uparrow$ & ACE (\%) $\downarrow$ & Time (s) $\downarrow$ \\
\midrule

MCP &  &  &58.40  &5.77  &0.09  & 50.06 & 6.30 & 0.12 & 73.81 & 3.71 & 0.04 \\
MCP + DE &  & \checkmark &60.24  &4.63  &0.48  & \textbf{51.80} & 5.01 & 0.67 & 74.21 & 2.90 & 0.26 \\
MCP + MC dropout &  & \checkmark &59.93  &4.71  &0.54  & 51.23 & 4.60 & 0.63 & 73.88 & 3.27 & 0.31 \\
\midrule

Temperature Scaling &  &  &58.40  &2.97  &0.11  & 50.06 & 3.06 & 0.15 & 73.81 & 2.66 & 0.06 \\
Temperature Scaling + DE &  & \checkmark &60.24  &2.21  &0.61  & \textbf{51.80} & 2.84 & 0.81 & 74.21 & 2.31 & 0.34 \\
Temperature Scaling + MC dropout &  & \checkmark  &59.93  &2.97  &0.67  & 51.23 & 2.26 & 0.73 & 73.88 & 2.23 & 0.40 \\
\midrule

logit-sampling  &\checkmark  &  &60.21  &2.11  &3.61  & 51.03 & 2.03 & 4.80 & 74.92 & 2.11 & 2.01 \\
logit-sampling + DE &\checkmark  & \checkmark &\textbf{60.54}  &1.83  &20.01  & 51.42 & \textbf{1.21} & 26.00 & \textbf{75.01} & \textcolor{blue}{1.78} & 12.01 \\
logit-sampling + MC dropout &\checkmark  & \checkmark &\textcolor{blue}{60.33}  &\textcolor{blue}{1.81}  &27.11  & \textcolor{blue}{51.70} & 1.63 & 29.60 & \textcolor{blue}{74.95} & 2.01 & 14.00 \\
\midrule

Our sampling-free approach & \checkmark &  &60.21  &2.06  &0.25  & 51.03 & 2.15 & 0.28 & 74.92 & 2.18 & 0.11 \\
Our sampling-free + DE & \checkmark & \checkmark &\textbf{60.54}  &\textbf{1.70}  &1.33  & 51.42 & \textcolor{blue}{1.33} & 1.46 & \textbf{75.01} & \textbf{1.73} & 0.73 \\
Our sampling-free + MC dropout & \checkmark & \checkmark &\textcolor{blue}{60.33}  &1.95  &1.48  & \textcolor{blue}{51.70} & 1.91 & 1.90 & \textcolor{blue}{74.95} & 1.94 & 0.88 \\
\midrule

EDL &\checkmark  &\checkmark  &57.33  &5.83  &0.18  & 48.30 & 5.32 & 0.15 & 68.01 & 2.91 & 0.09 \\
\bottomrule

\end{tabular}%
}
\end{table}

\begin{table}[ht!]
\centering
\caption{Comparison of inference time and FLOPs between our proposed sampling-free approach and logit-sampling for aleatoric uncertainty consideration in confidence estimation on the validation set of SemanticKITTI.}
\label{tab:flops_time}
\resizebox{0.6\linewidth}{!}{%
\begin{tabular}{lcc|cc|}
\toprule
\multirow{2}{*}{Method} & \multicolumn{2}{c|}{SalsaNext} & \multicolumn{2}{c}{RangeViT} \\
\cmidrule(lr){2-3} \cmidrule(lr){4-5}
& \textbf{Inference time} & \textbf{FLOPs} & \textbf{Inference time} & \textbf{FLOPs} \\
\midrule
Original model    & 0.12 ms & 62.62G & 0.09 ms & 52.01G\\

+ logit-sampling approach     & 4.80 ms & 69.44G & 3.61 ms & 58.83G\\
+ our sampling-free approach  & 0.28 ms & 62.69G & 0.25 ms & 52.08G \\
\bottomrule
\end{tabular}%
}
\end{table}


\section{Conclusion}
\label{sec:Conclusion}

We have developed a method to estimate the likelihood that the predicted class is correct, by examining the distributions of all possible class outcomes. We validated our sampling-free confidence estimation method on public datasets for LiDAR scene semantic segmentation, a field where safety-critical responses and real-time processing for large-scale data are crucial. Our comprehensive analysis comparing the lower bound confidences with the exact ones approximated through Monte Carlo integration, demonstrates a negligible discrepancy, confirming the robustness of our sampling-free lower bound confidence calibration approach for practical applications. Furthermore, when compared to the baseline approaches during inference, our proposed method consistently generates well-calibrated confidence values, exhibiting low ACE, calibrated reliability diagrams and fast inference. Moreover, our proposed method tends to be slightly underconfident across a broader range of regions. In conclusion, our proposed approach effectively performs semantic segmentation, ensuring accurate classification, well-calibrated confidence computation, and efficient performance, while also providing detailed uncertainty maps for pixel-wise semantic segmentation of LiDAR data.


\clearpage
\acknowledgments{If a paper is accepted, the final camera-ready version will (and probably should) include acknowledgments. All acknowledgments go at the end of the paper, including thanks to reviewers who gave useful comments, to colleagues who contributed to the ideas, and to funding agencies and corporate sponsors that provided financial support.}

\section{Limitations}
\label{sec:limitations}
Our results confirm that while our approach performs accurately and confidently on major classes well-represented during training—such as cars, roads, sidewalks, buildings, fences, and vegetation—it may struggle to distinguish between classes with similar features, occasionally confusing poles with thin trunks or bicycles with bicyclists. The proximity between Gaussian distributions does not necessarily result in misclassification if the predicted class maintains the highest mean. However, the closeness to another class's distribution increases predictive uncertainty, leading the model to be underconfident even when the object is correctly classified. 

Another limitation of this approach is the increase in training time compared to the original model, as it predicts the mean and variance to model Gaussian distributions over the logits for each class, which may require additional computational cost.

As a direction for future work, it would be valuable to investigate the use of normalizing flows as a more flexible alternative to directly predicting mean and variance for modeling class-conditional distributions. Additionally, evaluating the proposed approach for out-of-distribution detection and assessing its robustness under domain shift scenarios would further demonstrate its applicability to real-world settings.

\bibliography{references}  

\clearpage
\appendix
\section{Supplementary Material}
\label{sec:Supp}

\subsection{Derivation of the lower bound formula}
\label{sec:derivationLowerBound}
This derives the lower bound formula (Equation~\ref{eq:confidence}) of the main paper.

Given two Gaussians, $X\sim\mathcal{N}(x | \mu_a, \sigma_a^2) =: \varphi_a(x)$ and $Y\sim\mathcal{N}(\mu_b, \sigma_b^2)$, the probability $P(X > Y)$ is easily seen to be $P(X > Y) = \mathbb{E}_{\varphi_a}[\Phi_b(X)] = \Phi(\mu_a-\mu_b|0, \sigma^2_a + \sigma^2_b)$, where $\varphi(x)$ and $\Phi(x)$ denote the Gaussian PDF and CDF, respectively, and $\mathbb{E}_{\varphi_a}[\cdot]$ is the expectation over the distribution $\varphi_a$.

\begin{figure}[h]
\centering
\includegraphics[width=0.50\linewidth]{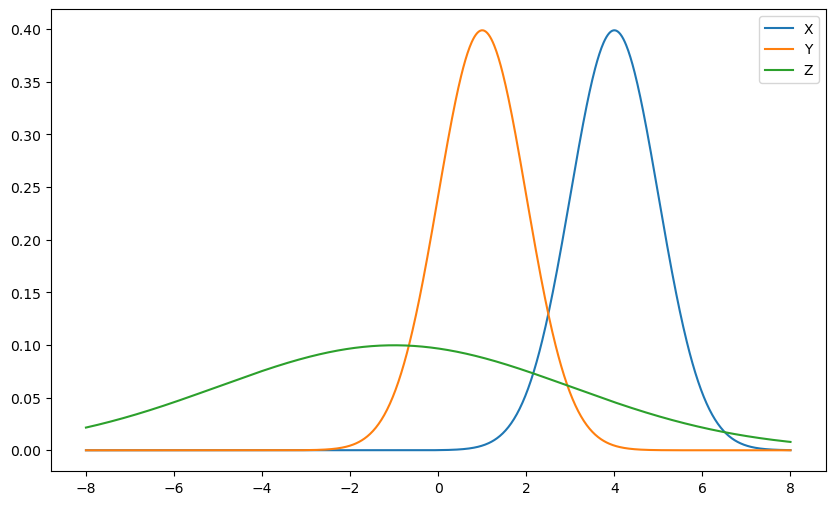}
  \caption{Illustration for three classes $X\sim \mathcal{N}(\mu_a=4, \sigma_a^2=1)$, $Y\sim \mathcal{N}(\mu_b=1, \sigma_b^2=1)$, and $Z\sim \mathcal{N}(\mu_c=-1, \sigma_c^2=4^2)$. As $\mu_a > \mu_b, \mu_c$, class A (rightmost peak) will be the predicted class. The pairwise confidences are $P(X>Y)=0.9831$ and $P(X>Z)=0.8874$ (so although $\mu_c<\mu_b$, it is more likely to confuse A with C than A with B, due to the large $\sigma_c$). The lower bound is $P(X>Y)\cdot P(X>Z)=0.8723$, whereas the exact value is $P(X>Y, Z)=0.8740 \geq 0.8723$, as expected.}
  \label{fig:three_gaussians}
\end{figure}

For three Gaussians (see Figure~\ref{fig:three_gaussians}), we are interested in $P(X > Y, Z) = P(X > \max(Y, Z)) = \mathbb{E}_{\varphi_a}[\Phi_b(X) \Phi_c(X)]$, for which there exists no closed-form solution. However, given the pairwise probabilities $P(X > Y)$ and $P(X > Z)$, their product is a lower bound, i.e., $\mathbb{E}_{\varphi_a}[\Phi_b(X) \Phi_c(X)] \geq \mathbb{E}_{\varphi_a}[\Phi_b(X)]\cdot \mathbb{E}_{\varphi_a}[\Phi_c(X)]$, as stated in Equation~\ref{eq:confidence} of the main paper and shown in the following.

In general, if $f$ and $g$ are strictly monotonically increasing functions over their full domain, then for a random variable $X$, the covariance of $f(X)$ and $g(X)$ will be non-negative. This is intuitively clear, since due to the monotonicity of the functions and their inverses, increasing or decreasing $f(X)$ will imply increasing or decreasing $g(X)$.

To prove, for the covariance of two random variables $Y$, $Z$, it generally holds that
\begin{align*}
\text{cov}(Y, Z) &\triangleq \mathbb{E}[(Y - \mathbb{E}[Y])\cdot(Z - \mathbb{E}[Z])] \\
&= \mathbb{E}[(Y - \mathbb{E}[Y])\cdot Z] \\
&= \mathbb{E}[(Y - \mathbb{E}[Y])\cdot(Z - a)]
\end{align*}
for any constant $a$. Setting $Y=f(X)$, $Z=g(X)$, and $a = g(f^{-1}(\mathbb{E}[f(X)]))$, we get:
\begin{align*}
& \text{cov}(f(X), g(X)) \\
&\triangleq  \mathbb{E}\left[ (f(X) - E[f(X)])\cdot (g(X) - \mathbb{E}[g(X)]) \right] \\
&=  \mathbb{E}\left[ (f(X) - \mathbb{E}[f(X)]) \cdot (g(X) - g(f^{-1}(\mathbb{E}[f(X)]))) \right].
\end{align*}
Since $f(X)$ is strictly monotonically increasing, the first term $f(X) - \mathbb{E}[f(X)]$ is positive if $X > f^{-1}(\mathbb{E}[f(X)])$ (due to strict monotonicity, $f^{-1}(y)$ is unique), and by construction, this also holds for the second term $g(X) - g(f^{-1}(\mathbb{E}[f(X)]))$, so that their product is always non-negative. It follows that $\text{cov}(f(X), g(X)) \geq 0$ and thus since in general, $\text{cov}(Y, Z) = \mathbb{E}[YZ] - \mathbb{E}[Y]\mathbb{E}[Z]$, it follows that $\mathbb{E}[YZ] \geq \mathbb{E}[Y]E[Z]$.

As $\Phi(x)$ are Gaussian CDFs, they are strictly monotonically increasing over the domain $x\in (-\inf, +\inf)$, and thus by setting $f(x)=\Phi_b(x)$, $g(x)=\Phi_c(x)$ and taking the expectation over $\varphi_a$, we obtain the claimed result $\mathbb{E}_{\varphi_a}[\Phi_b(X) \Phi_c(X)] \geq \mathbb{E}_{\varphi_a}[\Phi_b(X)]\cdot \mathbb{E}_{\varphi_a}[\Phi_c(X)]$.

For more than three Gaussians, the result is obtained recursively, noting that the product of any two Gaussian CDFs is also strictly monotonically increasing (which follows from the product rule and $\Phi(x)>0$ for all $x\in\mathbb{R}$).

Please note that the fact that we are computing $P(X > Y, Z)$ does not imply any limitation of our algorithm. Especially, there is no similarity to cases where a `one versus all' approach is used instead of `all versus all', leading to sub-optimal results (as in support vector machines). In our case, the winner class is determined by having the largest $\mu$, and we are only interested in the confidence associated with picking this winner class, which subsequently results in pairwise computations.

\subsection{Uncertainty-aware LiDAR semantic segmentation: qualitative analysis} 
\label{sec:LiDAR qualitative results}

This section presents qualitative evaluations of uncertainty maps generated by our sampling-free method combined with deep ensembles—the configuration achieving the lowest calibration error while maintaining competitive mIoU and fast inference. Overall, we observe high uncertainty not only at misclassified points but also along class boundaries (e.g., sidewalk–street transitions), beneath vehicles—where it is often ambiguous whether to label regions as vehicle or ground, even in manual annotations—around tree trunks, and in distant areas where LiDAR measurements become sparse and noisy. These observations demonstrate the effectiveness of our method in producing reliable uncertainty estimates, which are critical for safety-sensitive applications such as autonomous driving.
 
\begin{figure}[ht!]
\centering
    \begin{subfigure}{0.90\linewidth}
    \includegraphics[width=\linewidth]{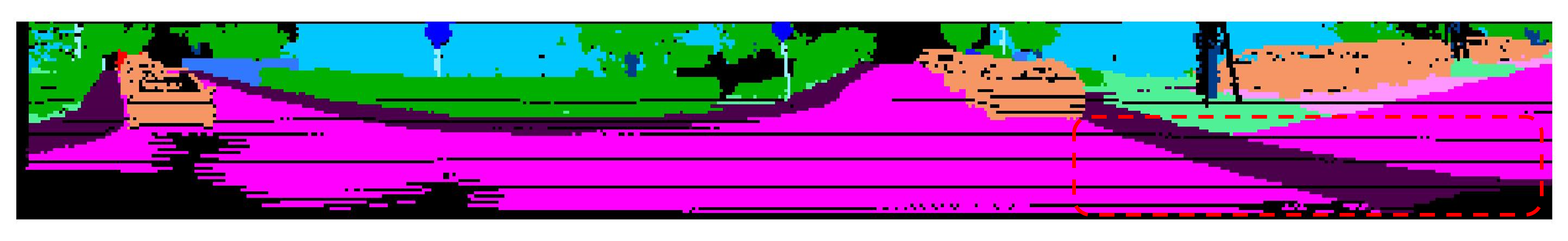}
    \caption{Ground truth.}
    \label{fig:Results_1-a}
    \end{subfigure}
  
    \begin{subfigure}{0.90\linewidth}
    \includegraphics[width=\linewidth]{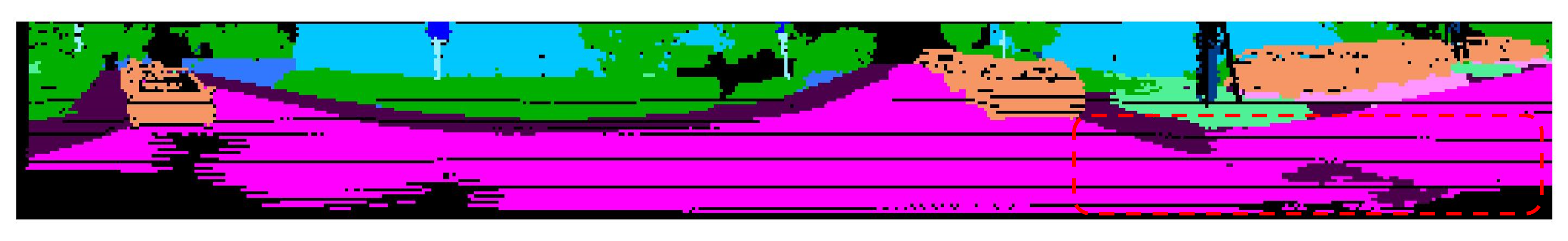}
    \caption{Our prediction.}
    \label{fig:Results_1-b}
    \end{subfigure}

    \begin{subfigure}{0.90\linewidth}
    \includegraphics[width=\linewidth]{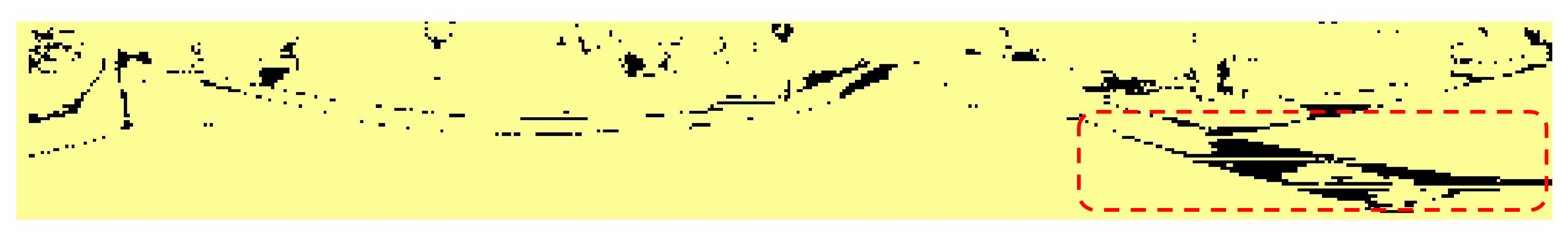}
    \caption{Our error map. \colorbox{CorrectClass}{correct classifications}, \colorbox{black}{\textcolor{white}{wrong classifications}}.}
    \label{fig:Results_1-c}
    \end{subfigure}
    
    \begin{subfigure}{0.90\linewidth}
    \includegraphics[width=\linewidth]{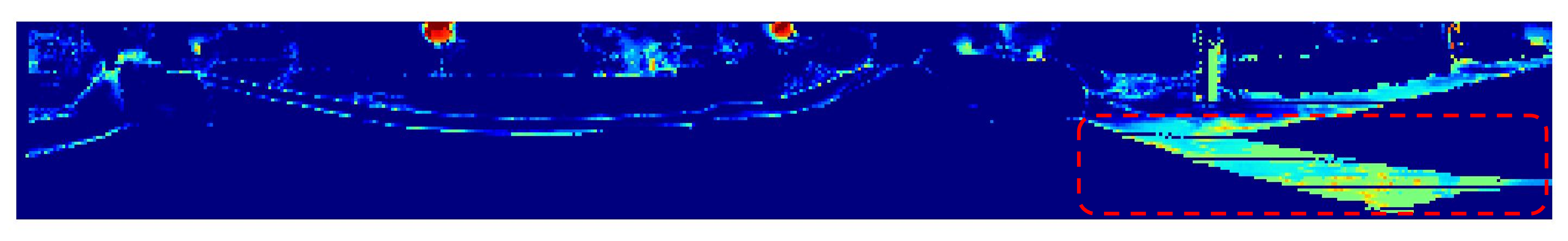}
    \caption{Our uncertainty map, from \colorbox{blue}{\textcolor{white}{low}} to \colorbox{red}{\textcolor{white}{high}} uncertainty.}
    \label{fig:Results_1-d}
    \end{subfigure}

    \begin{subfigure}{0.90\linewidth}
    \includegraphics[width=\linewidth]{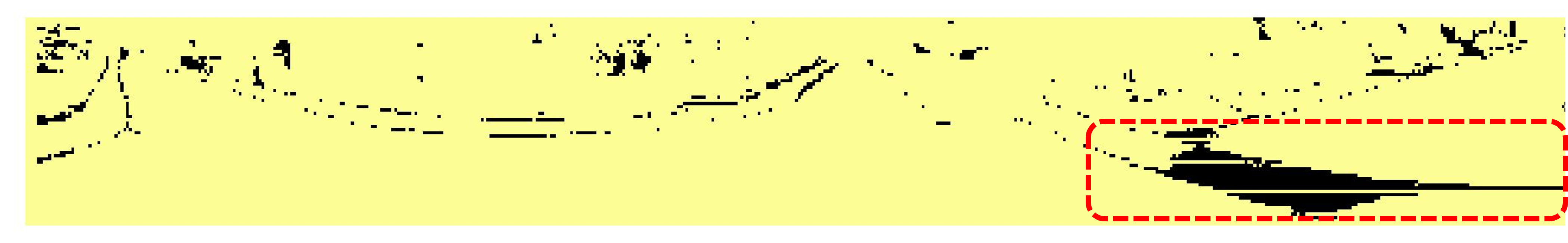}
    \caption{Error map for temperature scaling. \colorbox{CorrectClass}{correct classifications}, \colorbox{black}{\textcolor{white}{wrong classifications}}.}
    \label{fig:Results_1-e}
    \end{subfigure}
    
     \begin{subfigure}{0.90\linewidth}
    \includegraphics[width=\linewidth]{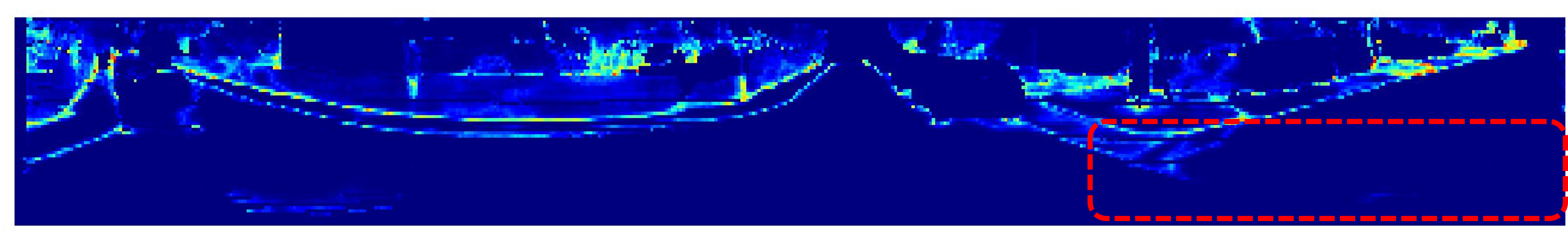}
    \caption{Uncertainty map for temperature scaling, from \colorbox{blue}{\textcolor{white}{low}} to \colorbox{red}{\textcolor{white}{high}} uncertainty.}
    \label{fig:Results_1-f}
    \end{subfigure}
    
    \begin{subfigure}{0.7\linewidth}
    \includegraphics[width=\linewidth]{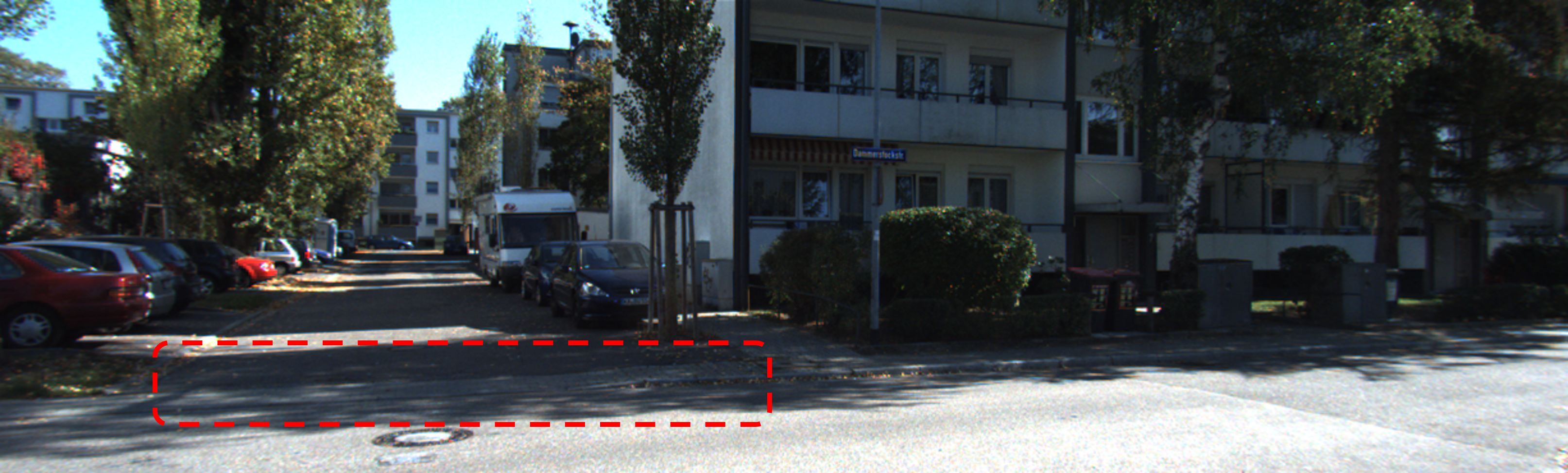}
    \caption{Camera image.}
    \label{fig:Results_1-g}
    \end{subfigure}
  
    \caption{Example of significant uncertainty arising from the confusion between the sidewalk and the street. The misclassified region (dashed red box) is labeled as a sidewalk in the ground truth but is also traversed by vehicles, causing overlapping classifications of street and sidewalk, which results in high uncertainty. Classes are represented with corresponding colors: \colorbox{outlier}{\textcolor{white}{outlier}},\colorbox{parking}{\textcolor{white}{parking}},\colorbox{car}{\textcolor{white} {car}},\colorbox{road}{\textcolor{white}{road}},\colorbox{sidewalk}{\textcolor{white}{sidewalk}},\colorbox{building}{\textcolor{white}{building}},\colorbox{fence}{\textcolor{white}{fence}},\colorbox{vegetation}{\textcolor{white}{vegetation}},\colorbox{trunk}{\textcolor{white}{trunk}},\colorbox{terrain}{terrain},\colorbox{pole}{pole},\colorbox{trafficsign}{\textcolor{white}{traffic-sign}}.}
    \label{fig:Results_1}
\end{figure}

Figure~\ref{fig:Results_1} presents the predicted segmentation, corresponding uncertainty map, and error map for a representative LiDAR scan. A region enclosed by a dashed red box—labeled as sidewalk in the ground truth—is ambiguously classified as street, likely because it is also accessible to vehicles in this area, as seen in the camera image (Figure~\ref{fig:Results_1-e}). However, for safety reasons, it is important to distinguish this shared area between pedestrians and cars from the normal street. This semantic ambiguity is effectively captured by our method, which assigns high uncertainty in this region. In contrast, the model calibrated with temperature scaling fails to reflect this ambiguity, assigning no uncertainty despite the misclassification (Figure~\ref{fig:Results_1-f}).

\begin{figure}[ht!]
\centering
    \begin{subfigure}{0.90\linewidth}
    \includegraphics[width=\linewidth]{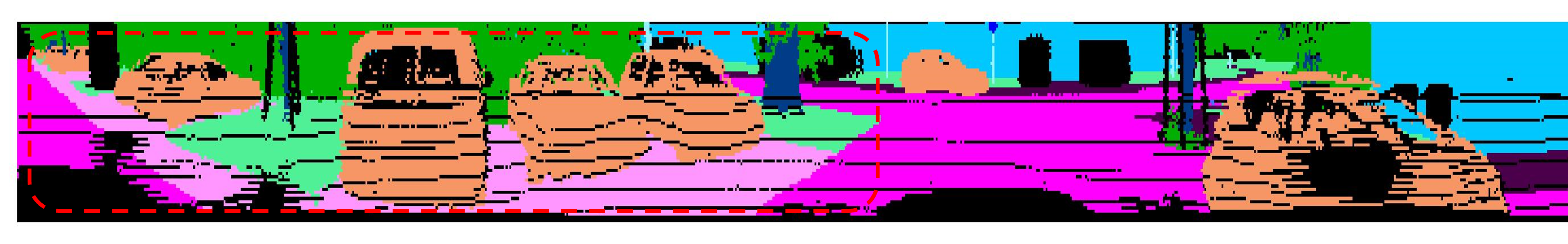}
    \caption{Ground truth.}
    \label{fig:Results_2-a}
    \end{subfigure}
    
    \begin{subfigure}{0.90\linewidth}
    \includegraphics[width=\linewidth]{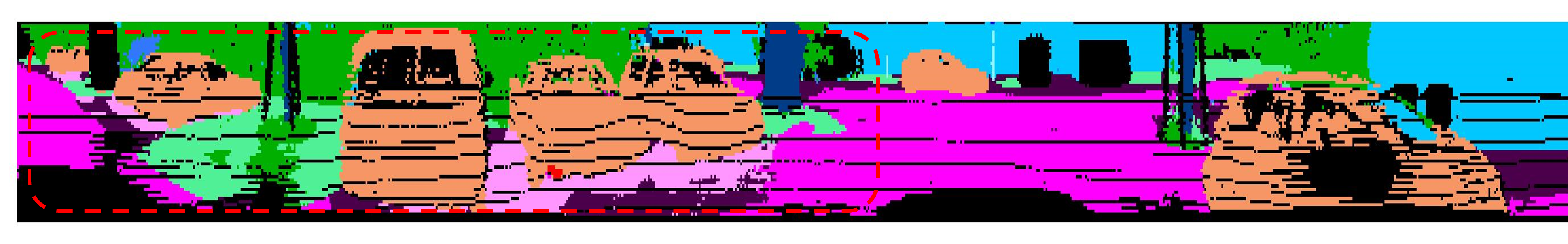}
    \caption{Prediction.}
    \label{fig:Results_2-b}
    \end{subfigure}
    
    \begin{subfigure}{0.90\linewidth}
    \includegraphics[width=\linewidth]{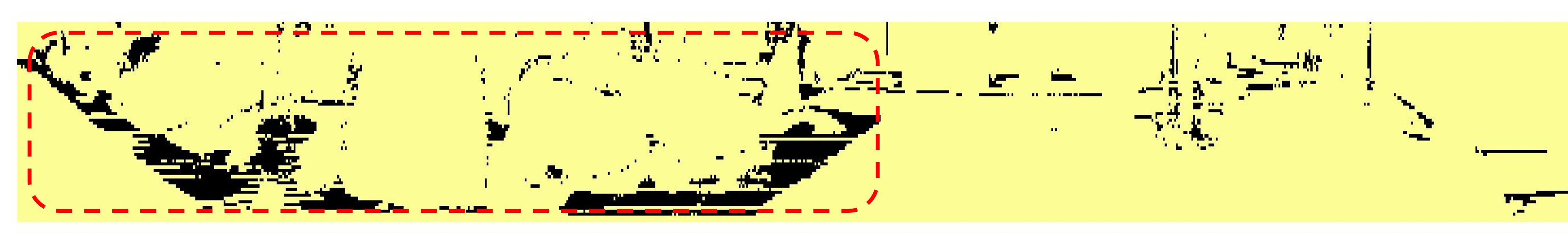}
    \caption{Error map. \colorbox{CorrectClass}{correct classifications}, \colorbox{black}{\textcolor{white}{wrong classifications}}.}
    \label{fig:Results_2-c}
    \end{subfigure}
    
    \begin{subfigure}{0.90\linewidth}
    \includegraphics[width=\linewidth]{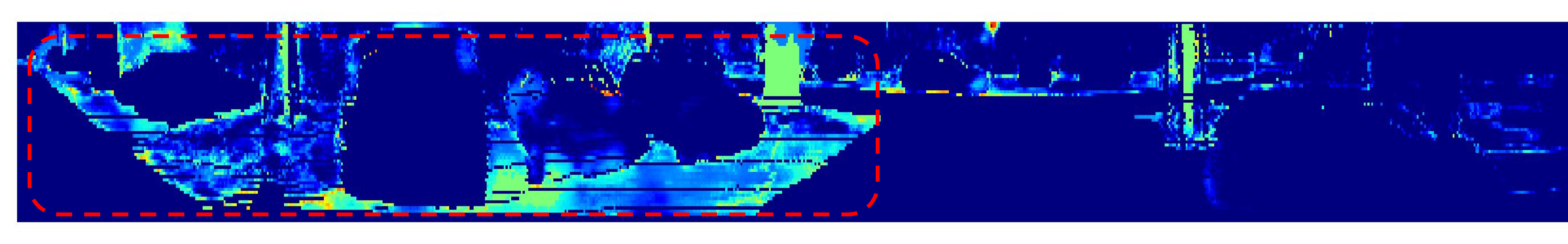}
    \caption{Uncertainty map, from \colorbox{blue}{\textcolor{white}{low}} to \colorbox{red}{\textcolor{white}{high}} uncertainty.}
    \label{fig:Results_2-d}
    \end{subfigure}
    
    \begin{subfigure}{0.7\linewidth}
    \includegraphics[width=\linewidth]{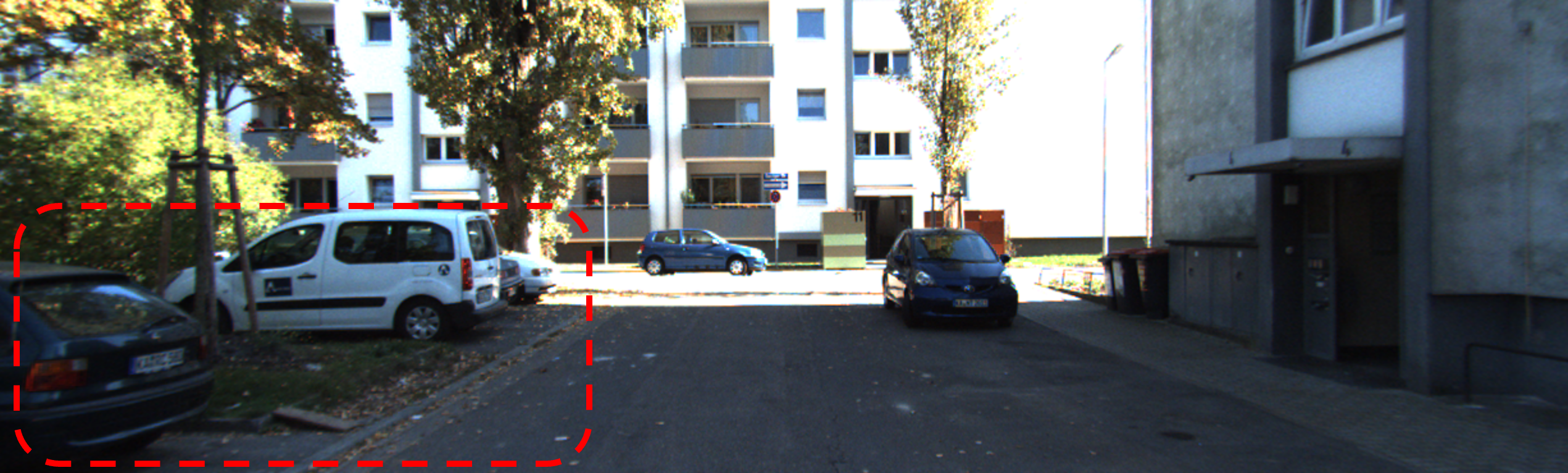}
    \caption{Camera image.}
    \label{fig:Results_2-e}
    \end{subfigure}
  
    \caption{Example of a misclassified ground region with high uncertainty. Uncertainty map highlights classification ambiguity, showing high uncertainty in regions with unclear ground class (e.g., terrain, sidewalk, or parking) due to overlapping Gaussian distributions. Classes are represented with corresponding colors: \colorbox{outlier}{\textcolor{white}{outlier}},\colorbox{parking}{\textcolor{white}{parking}},\colorbox{car}{\textcolor{white} {car}},\colorbox{road}{\textcolor{white}{road}},\colorbox{sidewalk}{\textcolor{white}{sidewalk}},\colorbox{building}{\textcolor{white}{building}},\colorbox{fence}{\textcolor{white}{fence}},\colorbox{vegetation}{\textcolor{white}{vegetation}},\colorbox{trunk}{\textcolor{white}{trunk}},\colorbox{terrain}{terrain},\colorbox{pole}{pole},\colorbox{trafficsign}{\textcolor{white}{traffic-sign}}.}

    \label{fig:Results_2}
\end{figure}

Another example, shown in Figure~\ref{fig:Results_2}, highlights a common challenge in autonomous driving: accurately identifying parking areas. In this case, the model exhibits uncertainty among three classes—road, terrain, and parking—and assigns high uncertainty to the region, effectively signaling that the classification in this area is unreliable.

\begin{figure*}[ht!]
\centering
    \begin{subfigure}{0.70\linewidth}
    \includegraphics[width=\linewidth]{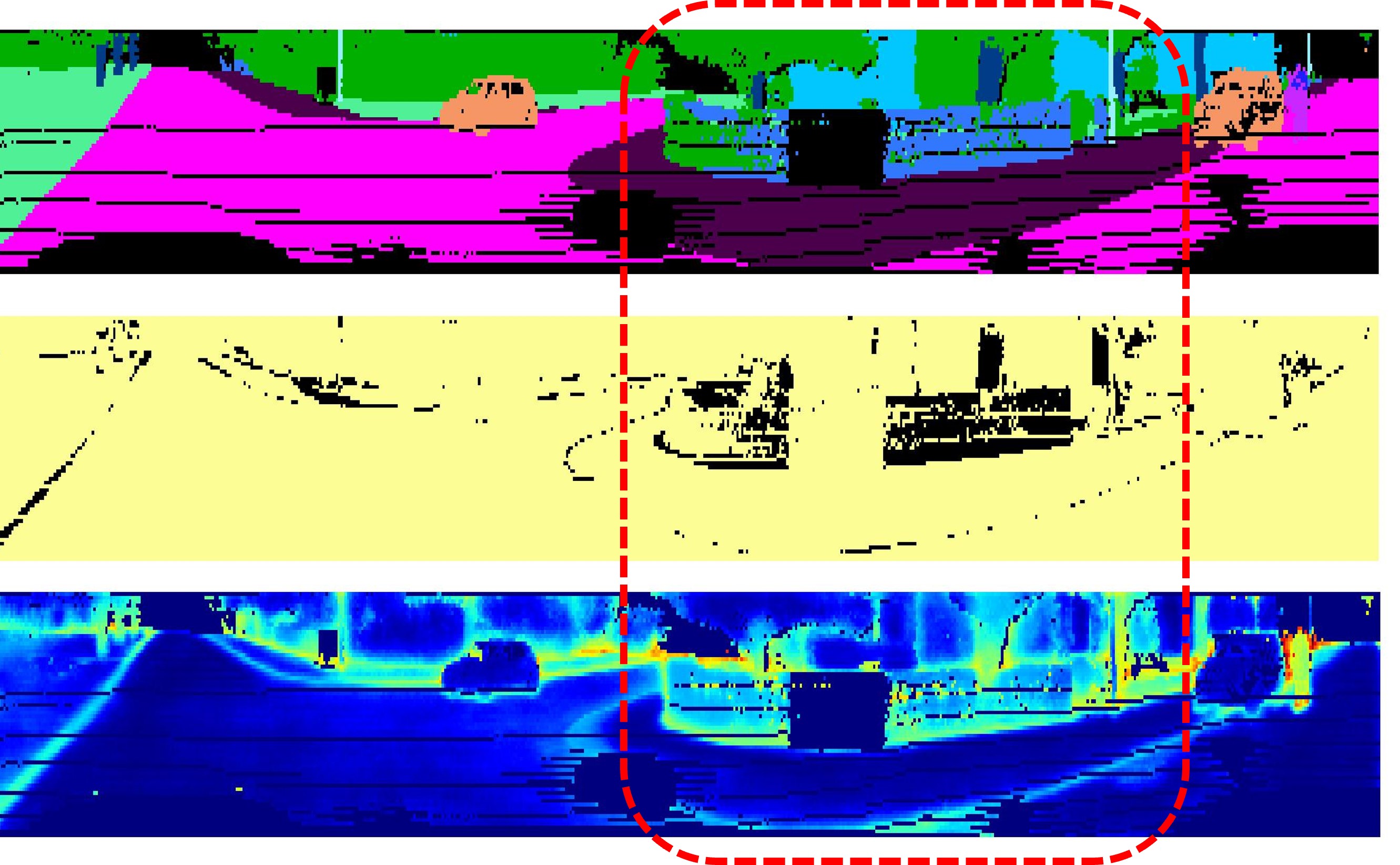}
    \caption{Predicted segmentation map, corresponding error map, and uncertainty map from our approach.}
    \label{fig:Results_3-a}
    \end{subfigure}
    
    \begin{subfigure}{0.70\linewidth}
    \includegraphics[width=\linewidth]{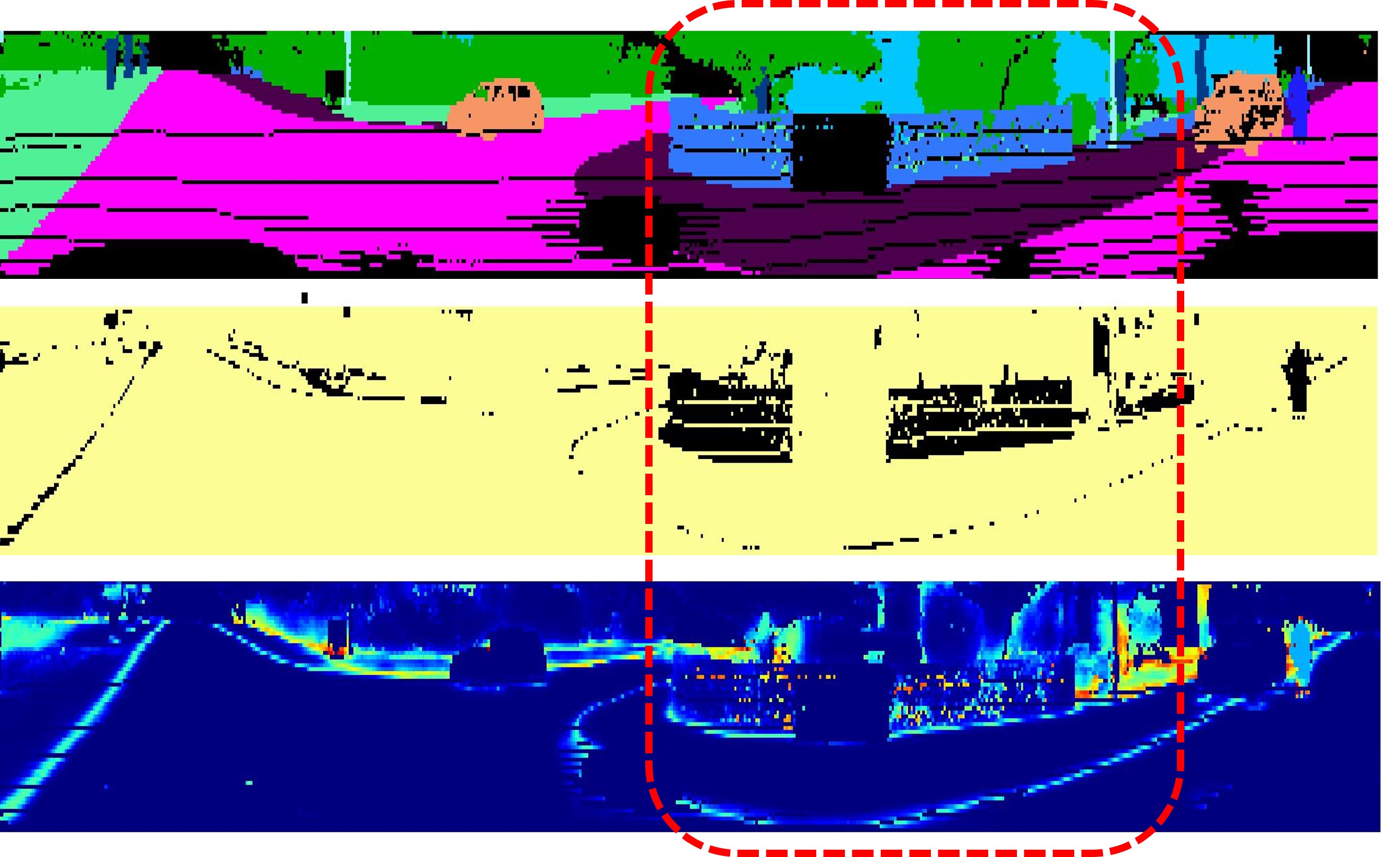}
    \caption{Predicted segmentation map, corresponding error map, and uncertainty map from temperature scaling.}
    \label{fig:Results_3-b}
    \end{subfigure}
    
    \begin{subfigure}{0.70\linewidth}
    \includegraphics[width=\linewidth]{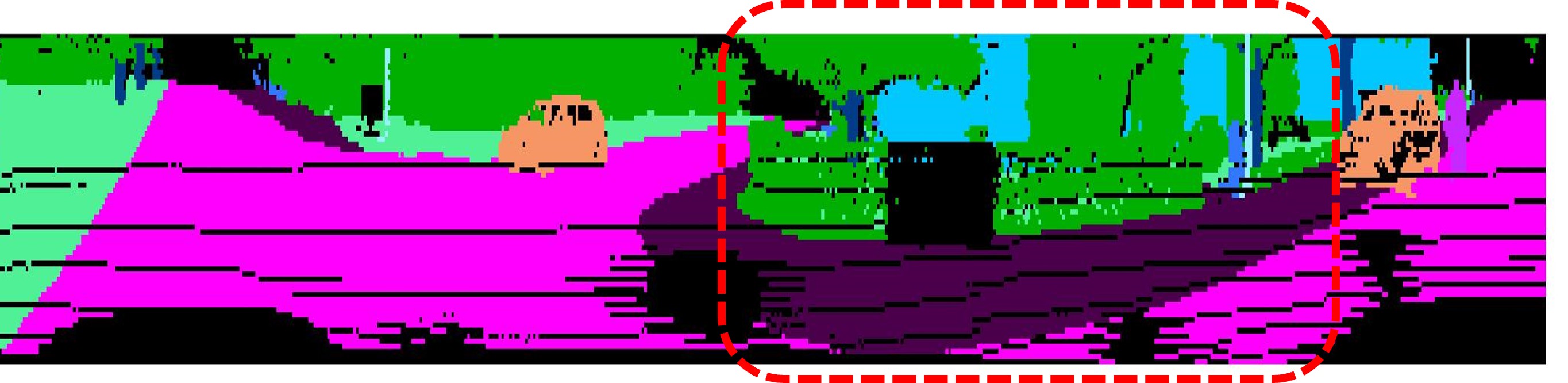}
    \caption{Ground truth.}
    \label{fig:Results_3-c}
    \end{subfigure}
    
    \begin{subfigure}{0.60\linewidth}
    \includegraphics[width=\linewidth]{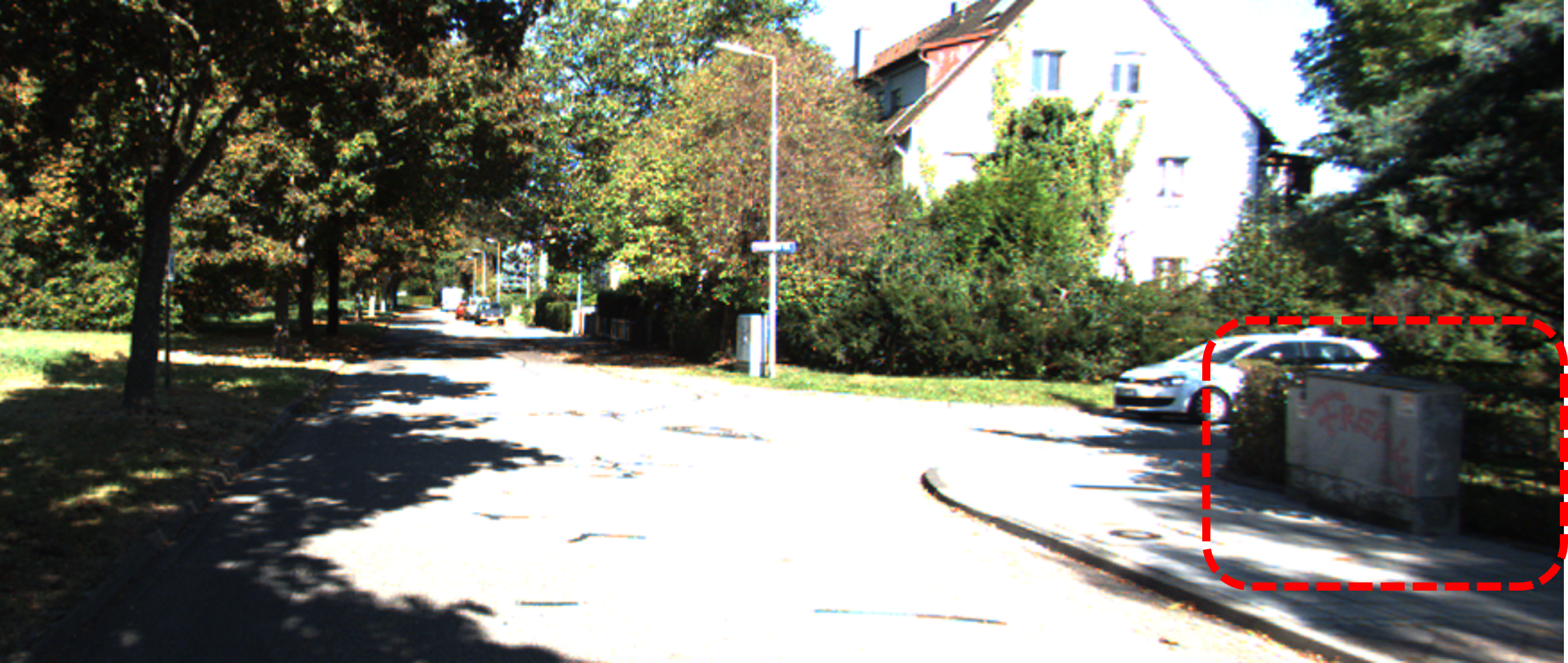}
    \caption{Camera image.}
    \label{fig:Results_3-d}
    \end{subfigure}

    \caption{Comparison of uncertainty estimates for a semantically ambiguous region appearing as a mixture of fence and vegetation in the camera image (\ref{fig:Results_3-d}) but labeled solely as vegetation in the ground truth (\ref{fig:Results_3-c}). Our approach (\ref{fig:Results_3-a}) effectively assigns high uncertainty to the entire region, while temperature scaling (\ref{fig:Results_3-b}) underestimates uncertainty, predicting only a few points with high uncertainty. Classes are represented with corresponding colors: \colorbox{outlier}{\textcolor{white}{outlier}},\colorbox{parking}{\textcolor{white}{parking}},\colorbox{car}{\textcolor{white} {car}},\colorbox{road}{\textcolor{white}{road}},\colorbox{sidewalk}{\textcolor{white}{sidewalk}},\colorbox{building}{\textcolor{white}{building}},\colorbox{fence}{\textcolor{white}{fence}},\colorbox{vegetation}{\textcolor{white}{vegetation}},\colorbox{trunk}{\textcolor{white}{trunk}},\colorbox{terrain}{terrain},\colorbox{pole}{pole},\colorbox{trafficsign}{\textcolor{white}{traffic-sign}}.}

    \label{fig:Results_3}
\end{figure*}

Figure~\ref{fig:Results_3} highlights an uncertain region in the dashed red box that appears in the camera image(\ref{fig:Results_3-d}) as a mixture of fence and vegetation, but is labeled solely as vegetation in the ground truth (\ref{fig:Results_3-c}). Our proposed approach (\ref{fig:Results_3-a}) correctly identifies this area as highly uncertain, reflecting the semantic ambiguity. In contrast, the temperature scaling method (\ref{fig:Results_3-b}) assigns low uncertainty to most of the region, with only a few isolated points marked as uncertain, despite the overall unreliability of the classification. This example also shows classes with low uncertainty, such as street, cars and sidewalk, which are classified correctly and achieving low uncertainty prediction. 

\subsection{Evaluation of our approach on classification tasks} 
\label{sec:classification}
We evaluated our experiments on CIFAR-10 and CIFAR-100 \citep{krizhevsky2009learning} as benchmark datasets for classification, each containing $60,000$ $[32 \times 32 \times 3]$ color images, with CIFAR-10 divided into 10 classes and CIFAR-100 into $100$ classes, providing a robust test of our approach with a varying number of classes. For these classification tasks, we utilized VGG-16 and Wide-ResNet-28-10 architectures \citep{simonyan2014very}.

\begin{table*}[ht!]
\centering
\caption{Comparative analysis of inference time, mIoU, and ACE across various confidence calibration approaches for classification tasks on CIFAR-10 and CIFAR-100 using both the VGG-16 and WideResNet models, and for semantic segmentation on SemanticKITTI using the SalsaNext model.} 

\label{tab:time_ace}
\resizebox{\linewidth}{!}{%
\begin{tabular}{@{}lccccccccccccccc@{}}
\toprule
& \multicolumn{3}{c}{CIFAR-10, VGG-16} & \multicolumn{3}{c}{CIFAR-10, Wide-ResNet-28-10} & \multicolumn{3}{c}{CIFAR-100, VGG-16} & \multicolumn{3}{c}{CIFAR-100, Wide-ResNet-28-10} & \multicolumn{3}{c}{SemanticKITTI} \\ 
\cmidrule(lr){2-4} \cmidrule(lr){5-7} \cmidrule(lr){8-10} \cmidrule(lr){11-13} \cmidrule(lr){14-16}
Method & Accuracy(\%) $\uparrow$ & ACE (\%) $\downarrow$ & Time (s) $\downarrow$ & Accuracy(\%) $\uparrow$ & ACE (\%) $\downarrow$ & Time (s) $\downarrow$ & Accuracy(\%) $\uparrow$ & ACE (\%) $\downarrow$ & Time (s) $\downarrow$ & Accuracy(\%) $\uparrow$ & ACE (\%) $\downarrow$ & Time (s) $\downarrow$ & mIoU(\%) $\uparrow$ & ACE (\%) $\downarrow$ & Time (s) $\downarrow$ \\
\midrule

MCP  &93.40  &8.66 &0.02      &95.12 &5.21 &0.03     &72.81 &9.01 &0.08     &79.01 &6.06 &0.13    &52.06  &6.81 &0.12 \\
MCP + DE &94.01 &6.89 &0.13      &95.70 &3.38 &0.18     &75.68 &7.86 &0.50     &80.23 &5.43 &0.84    &\textbf{53.80} &5.06 & 0.67\\
MCP + MC dropout  &93.47 &6.71 &0.21      &95.87 &3.30 &0.23     &75.01 &7.40 &0.73     &80.74 &5.21 &1.20    &53.03 &4.60 & 0.63 \\
\midrule

logit-sampling (50 samples) &93.78   &5.16 &0.25     &95.07 &2.07 &0.31     &73.44 &6.74 &1.80     &80.13 &4.31 &2.61    &50.89 & 5.30 &4.80  \\
logit-sampling (50 samples)+DE & 94.40 & 1.66 & 1.32   &96.50 &1.23 &1.50     &75.07 &\textbf{1.16} &10.00  &81.46 &4.01 &12.08    & 51.42 & \textbf{2.95} & 26.00 \\
logit-sampling (50 samples)+MC dropout & 93.93 & 1.40 & 2.70   &96.10 &1.20 &2.20     &76.50 &1.23 &10.80     &81.08 &3.98 &12.28    &51.70 &3.10 &29.60 \\
\midrule

Our sampling-free approach &93.10  &5.76 & 0.03     &95.60 &1.91 &0.06     &73.44 &6.98 & 0.24    &80.10 &4.09 &0.37    &51.03 &4.90 &0.28\\
Our sampling-free approach+DE &\textbf{94.81} &\textbf{1.21} &0.17    &\textbf{96.66} &\textbf{0.81} &0.31     &\textbf{77.03} &1.80 &1.45      &\textbf{82.16} &\textbf{3.61} &1.99    & 51.60  &\textbf{3.30} &1.46 \\
Our sampling-free approach+MC dropout &93.91 &1.40 &0.33   &95.82 &0.89 &0.47     &75.02 &1.43 &1.45     &81.70 &3.89 &2.20    &52.20 &3.01 &1.90  \\
\midrule

Temperature Scaling     &93.40 &1.60 &0.03     &95.12 &2.40 &0.05     &72.81 &2.08 &0.10     &79.01 &4.68 &0.18    &52.06 &4.29 &0.15  \\
EDL   &92.02 &4.30 & 0.06     &94.70 &3.66 &0.11     &73.40 &5.26 &0.10     &78.50 &5.01 &0.23    & 48.30 & 5.32 & 0.15 \\
\bottomrule

\end{tabular}%
}

\end{table*}

\end{document}